\documentclass[10pt]{article} % For LaTeX2e
\usepackage[preprint]{tmlr}

\usepackage{url}
\usepackage{tikz}
\usepackage{comment}
\usepackage{amsmath,amssymb} % define this before the line numbering.
\usepackage{color}
\usepackage[ruled,vlined]{algorithm2e}

\usepackage[accsupp]{axessibility}  % Improves PDF readability for those with disabilities.

\usepackage[skip=3pt]{caption}

\DeclareCaptionFormat{myformat}{\fontsize{9}{9}\selectfont#1#2#3}
\usepackage{graphicx}
\usepackage{amsmath}
\usepackage{amssymb}
\usepackage{booktabs}

\usepackage{amsfonts}
\usepackage{subcaption}

\usepackage{url}
\usepackage{wrapfig}
\usepackage{verbatim}
\usepackage{graphicx}
\usepackage{caption}
\usepackage{wrapfig}
\usepackage{diagbox}
\usepackage{multirow}
\usepackage{color}
\definecolor{myred}{rgb}{1,0,0} 
\usepackage{enumitem}
\usepackage[toc, page]{appendix}
\usepackage{xcolor} 
\usepackage{url}
\usepackage{color, colortbl}
\usepackage{esvect}
\usepackage{wrapfig}
\definecolor{LightCyan}{rgb}{0.88,1,1}
\definecolor{cyberpink}{RGB}{235,110, 152}
\definecolor{lbrown}{RGB}{230,100, 20}
\definecolor{lime}{RGB}{205, 237, 250}
\definecolor{pink}{RGB}{252, 237, 246}

\definecolor{rlime}{RGB}{205, 237, 250}
\definecolor{rlemon}{RGB}{255, 250, 186}

\definecolor{lemon}{RGB}{218, 243, 245}
\newcommand{\genie}[0]{\texttt{GeNIe}}
\newcommand{\genienoad}[0]{\texttt{GeNIe-Ada}}
\newcommand{\posl}[0]{\texttt{Img2Img}$^{L}$}
\newcommand{\posh}[0]{\texttt{Img2Img}$^{H}$}
\newcommand{\txtimg}[0]{\texttt{Txt2Img}}
\newcommand{\imgimg}[0]{\texttt{Img2Img}}

\DeclareMathOperator*{\argmaxA}{argmax}
\usepackage{enumitem}
\setlist{leftmargin=2.0mm}
\usepackage{graphicx}
\usepackage{booktabs}

\usepackage[accsupp]{axessibility}  % Improves PDF readability for those with disabilities.

\usepackage[colorlinks]{hyperref}
 \hypersetup{
     allcolors = lbrown
     }

\newcommand\blfootnote[1]{%
  \begingroup
  \renewcommand\thefootnote{}\footnote{#1}%
  \addtocounter{footnote}{-1}%
  \endgroup
}

\title{GeNIe: Generative Hard Negative Images Through Diffusion}

\begin{document}

\author{
\centering
\name Soroush Abbasi Koohpayegani$^{*,1}$ $\quad$  $\quad$ Anuj Singh$^{*,2,3}$ \\ K L Navaneet$^{1}$   $\quad$  $\quad$ Hamed Pirsiavash$^{1}$ $\quad$  $\quad$ Hadi Jamali-Rad$^{2,3}$ \\
\addr $^{1}$University of California, Davis \\
$^{2}$Delft University of Technology, The Netherlands\\ 
$^{3}$Shell Global Solutions International B.V., Amsterdam, The Netherlands \\
\{soroush,nkadur,hpirsiav\}@ucdavis.edu \quad \{a.r.singh,h.jamalirad\}@tudelft.nl
}

\maketitle

\begin{abstract}
Data augmentation is crucial in training deep models, preventing them from overfitting to limited data. Recent advances in generative AI, e.g., diffusion models, have enabled more sophisticated augmentation techniques that produce data resembling natural images. We introduce \genie{} a novel augmentation method which leverages a latent diffusion model conditioned on a text prompt to combine two contrasting data points (an image from the source category and a text prompt from the target category) to generate challenging augmentations. To achieve this, we adjust the noise level (equivalently, number of diffusion iterations) to ensure the generated image retains low-level and background features from the source image while representing the target category, resulting in a \emph{hard negative} sample for the source category. We further automate and enhance \genie{} by adaptively adjusting the noise level selection on a per image basis (coined as \texttt{GeNIe-Ada}), leading to further performance improvements. Our extensive experiments, in both few-shot and long-tail distribution settings, demonstrate the effectiveness of our novel augmentation method and its superior performance over the prior art. Our code is available at: \url{https://github.com/UCDvision/GeNIe}.
\end{abstract}

\blfootnote{* equal contribution}

\section{Introduction}
\label{sec:intro}

Augmentation has become an integral part of training deep learning models, particularly when faced with limited training data. For instance, when it comes to image classification with limited number of samples per class, model generalization ability can be significantly hindered. 
% Same goes with more complex tasks such as detection and segmentation in data-deficient settings.
Simple transformations like rotation, cropping, and adjustments in brightness artificially diversify the training set, offering the model a more comprehensive grasp of potential data variations. 
% Exposure to a broader range of augmented samples enhances model robustness, adaptability, and accuracy in predicting novel instances. 
Hence, augmentation can serve as a practical strategy to boost the model's learning capacity, minimizing the risk of overfitting and facilitating effective knowledge transfer from limited labelled data to real-world scenarios. Various image augmentation methods, encompassing standard transformations, and learning-based approaches have been proposed \cite{cubuk2019randaugment, cubuk2019autoaugment, Cutmix, mixup, trabucco2024effective}. Some augmentation strategies combine two images possibly from two different categories to generate a new sample image. The simplest ones in this category are MixUp \citep{mixup} and CutMix \citep{Cutmix} where two images are combined in the pixel space. 
However, the resulting augmentations often do not lie within the manifold of natural images and act as out-of-distribution samples that will not be encountered during testing.
% However, the resulting augmentations are usually not in the manifold of natural images, making it unlikely for the model to see such images at test time. %\hamed{ours is similar to this in using two separate source of info. Add it to abstract or conclusion.}

Recently, leveraging generative models for data augmentation has gained an upsurge of attention \citep{trabucco2024effective, roy2023cap2aug, luzi2022boomerang, he2022synthetic}. These interesting studies, either based on fine-tuning or prompt engineering of diffusion models, are mostly focused on generating \emph{generic augmentations} without considering the impact of other classes and incorporating that information into the generative process for a classification context. We take a different approach to generate challenging augmentations near the decision boundaries of a downstream classifier. Inspired by diffusion-based image editing methods \citep{meng2021sdedit, luzi2022boomerang} some of which are previously used for data augmentation, we propose to use conditional latent diffusion models \citep{rombach2021highresolution} for generating \emph{hard negative} images. Our core idea (coined as \genie) is to sample source images from various categories and prompt the diffusion model with a contradictory text corresponding to a different target category. We demonstrate that the choice of noise level (or equivalently number of iterations) for the diffusion process plays a pivotal role in generating images that semantically belong to the target category while retaining low-level features from the source image. We argue that these generated samples serve as \emph{hard negatives} \citep{xuan2021hard, mao2017help} for the source category (or from a dual perspective hard positives for the target category). To further enhance \genie, we propose an adaptive noise level selection strategy (dubbed as \texttt{GeNIe-Ada}) enabling it to adjust noise levels automatically per sample.

\begin{figure*}[t]
\centering
\includegraphics[width=1.0\linewidth]{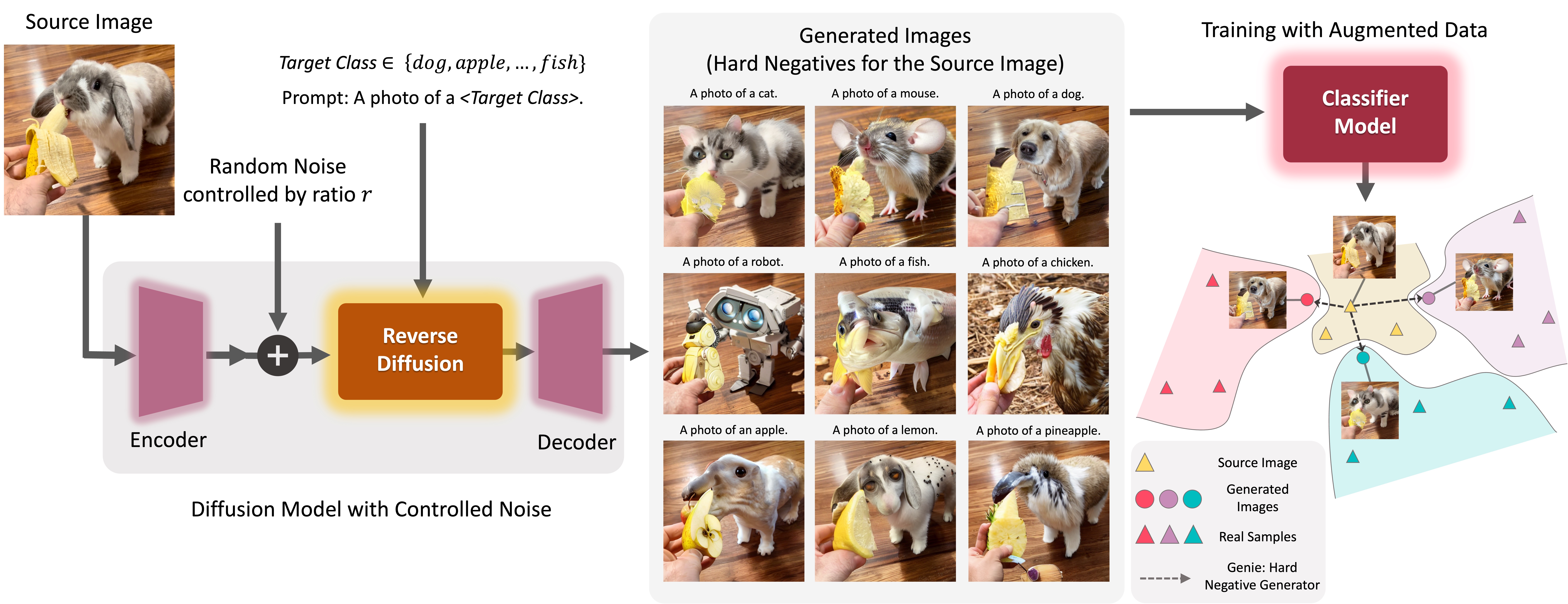}
\caption{{\bf \underline{Ge}nerative Hard \underline{N}egative \underline{I}mag\underline{e}s Through Diffusion (\genie):} generates hard negative images that belong to the target category but are similar to the source image from low-level feature and contextual perspectives. \genie{} starts from a source image passing it through a partial noise addition process, and conditioning it on a different target category. By controlling the amount of noise, the reverse latent diffusion process generates images that serve as \emph{hard negatives} for the source category.}
\label{fig:genie}
\vspace{0.3cm}
\end{figure*}

To establish the impact of \genie{}, we focus on two challenging scenarios: \emph{long-tail} and \emph{few-shot} settings. In real-world applications, data often follows a long-tail distribution, where common scenarios dominate and rare occurrences are underrepresented. For instance, a person jaywalking a highway causes models to struggle with such unusual scenarios. Combating such a bias or lack of sufficient data samples during model training is essential in building robust models for self-driving cars or surveillance systems, to name a few. Same challenge arises in few-shot learning settings where the model has to learn from only a handful of samples. Our extensive quantitative and qualitative experimentation, on a suite of few-shot and long-tail distribution settings, corroborate the effectiveness of the proposed novel augmentation method (\genie{}, \texttt{GeNIe-Ada}) in generating hard negatives, corroborating its significant impact on categories with a limited number of samples. A high-level sketch of \genie{} is illustrated in Fig.~\ref{fig:genie}. Our main contributions are summarized below: 
% As can be seen, \genie{} can take as input a source image of a bunny and generate images augmenting $9$ different target categories in which the low-level features and background context of the source image are preserved.  

\begin{itemize}
    \item[-] We introduce \genie{}, a novel yet elegantly simple diffusion-based augmentation method to create challenging augmentations in the manifold of natural images. For the first time, to our best knowledge, \genie{} achieves this by combining two sources of information (a source image, and a contradictory target prompt) through a noise-level adjustment mechanism.
    
    \item[-] We further extend \genie{} by automating the noise-level adjustment strategy on a per-sample basis (called \texttt{GeNIe-Ada}), to enable generating hard negative samples in the context of image classification, leading also to further performance enhancement. 

    \item[-] To substantiate the impact of \genie{}, we present a suit of quantitative and qualitative results including extensive experimentation on two challenging tasks: few-shot and long tail distribution settings corroborating that \genie{} (and its extension \texttt{GeNIe-Ada}) significantly improve the downstream classification performance.     
\end{itemize}

\section{Related Work}
% \vspace{-.1in}
\textbf{Data Augmentations.} Simple flipping, cropping, colour jittering, and blurring are some forms of image augmentations \citep{shorten2019survey}. These augmentations are commonly adopted in training deep learning models. However, using these data augmentations is not trivial in some domains. For example, using blurring might remove important low-level information from medical images. More advanced approaches, such as MixUp \citep{mixup} and CutMix \citep{Cutmix}, mix images and their labels accordingly \citep{hendrycks2020augmix, liu2022automix, kim2020puzzle, random_augment}. However, the resulting augmentations are not natural images anymore, and thus, act as out-of-distribution samples that will not be seen at test time. Another strand of research tailors the augmentation strategy through a learning process to fit the training data  \citep{ding2024saflex, cubuk2019randaugment, cubuk2019autoaugment}. Unlike the above methods, we propose to utilize pre-trained latent diffusion models to generate hard negatives (in contrast to generic augmentations) through a noise adaptation strategy discussed in Section \ref{sec:method}. 

%weak statement creatin unnecessary debate %Note that generative models are indeed trained to generate natural images; however, the typical challenge here is that the generated images might not necessarily belong to same data distribution as the training dataset, which could entail further fine-tuning them to the specific domain.

\textbf{Data Augmentation with Generative Models.} Using synthesized images from generative models to augment training data has been studied before in many domains \citep{gan_medical, gan_dom_adaptatino}, including domain adaptation \citep{aug_gan}, visual alignment \citep{peebles2022gansupervised}, and mitigation of dataset bias \citep{hendricks_tyranny, hemmat2023feedbackguided, prabhu2024lance}. For example, \citep{prabhu2024lance} introduces a methodology aimed at enhancing test set evaluation through augmentation. While previous methods predominantly relied on GANs \citep{zhang21, li2022bigdatasetgan, Tritrong2021RepurposeGANs} as the generative model, more recent studies promote using diffusion models to augment the data \citep{rombach2021highresolution, he2022synthetic, shipard2023boosting, trabucco2024effective, azizi2023synthetic, luo2023camdiff, roy2023cap2aug, jain2022distilling, feng2023diverse, dunlap2023diversify, chegini2023identifying}. More specifically, \citep{trabucco2024effective, roy2023cap2aug, he2022synthetic, azizi2023synthetic} study the effectiveness of text-to-image diffusion models in data augmentation by diversification of each class with synthetic images. \citep{trabucco2024effective} leverages a text-to-image diffusion model and fine-tunes it on the downstream dataset using textual-inversion \citep{gal2022image} to increase the diversity of existing samples. \citep{roy2023cap2aug} also utilizes a text-to-image diffusion model, but with a BLIP \citep{li2022blip} model to generate meaningful captions from the existing images. \citep{jain2022distilling} utilizes diffusion models for augmentation to correct model mistakes. \citep{feng2023diverse} uses CLIP \citep{radford2021learning} to filter generated images. \citep{dunlap2023diversify} utilizes text-based diffusion and a large language model (LLM) to diversify the training data. \citep{chegini2023identifying} uses an LLM to generate text descriptions of failure modes associated with spurious correlations, which are then used to generate synthetic data through generative models. The challenge here is that the LLM has little understanding of such failure scenarios and contexts. 

We take a completely different approach here, without replying on any extra source of information (e.g., through an LLM). Inspired by image editing approaches such as Boomerang \citep{luzi2022boomerang} and SDEdit \citep{meng2021sdedit}, we propose to adaptively guide a latent diffusion model to generate \emph{hard negatives} images \citep{mao2017help, xuan2021hard} on a per-sample basis per category. In a nutshell, the aforementioned studies focus on improving the diversity of each class with effective prompts and diffusion models, however, we focus on generating effective \emph{hard negative} samples for each class by combining two sources of contradicting information (images from the source category and text prompt from the target category).

% We build a structured approach (described in Section \ref{sec:method}), by taking a source sample image (for instance from a minority class) and propose an adaptive noise adjustment strategy next to a contradicting text prompt referring to a target class, so that the generated augmentation represents the target class while preserving the visual futures of the source samples, and thus acting as \emph{hard negative} for the source category. In a nutshell, the aforementioned studies focus on improving diversity of each class with effective prompts and diffusion models, however, we focus on generating effective \emph{hard negative} samples for each class by combining two sources of contradicting information (images from the source category and text prompt from the target category).

\textbf{Language Guided Recognition Models.} Vision-Language foundation models (VLMs) \citep{alayrac2022flamingo, radford2021learning, rombach2021highresolution, saharia2022photorealistic, ramesh2022hierarchical, ramesh2021zero} utilize human language to guide the generation of images or to extract features from images that are aligned with human language. 
% Due to alignment with human language, these models can be used in downstream recognition tasks. 
For example, CLIP \citep{radford2021learning} shows decent zero-shot performance on many downstream tasks by matching images to their text descriptions. Some recent works improve the utilization of human language in the prompt \citep{lads, gals}, and others use a diffusion model directly as a classifier \citep{li2023diffusion}. Similar to the above, we use a foundation model (Stable Diffusion 1.5 \citep{rombach2021highresolution}) to improve the downstream task. Concretely, we utilize category names of the downstream tasks to augment their associate training data with hard negative samples.

\textbf{Few-Shot Learning.} In Few-shot Learning (FSL), we pre-train a model with abundant data to learn a rich representation, then fine-tune it on new tasks with only a few available samples. In supervised FSL \citep{chen2019a, afrasiyabi2019associative, qiao2018few, ye2020few, dvornik2019diversity, li2020adversarial, sung2018learning, zhou2021binocular, singh2023transductive}, pretraining is done on a labeled dataset, whereas in unsupervised FSL \citep{psco,wang2022contrastive,unisiam, Qin2020ULDA, Antoniou2020Assume, khodadadeh2019unsupervised, hsu2018unsupervised, medina2020self, Shirekar_2023_WACV} the pretraining has to be conducted on an unlabeled dataset posing an extra challenge in the learning paradigm and neighboring these methods closer to the realm of self-supervised learning. Even though FSL is not of primal interest in this work, we assess the impact of \texttt{GeNIe} on a number of few-shot scenarios and state-of-the-art baselines by accentuating on its impact on the few-shot inference stage.

% revolves around learning very fast from only a handful of samples. FSL is typically conducted in two stages: pretraining on an abundance of data followed by fast adaptation to unseen few-shot episodes. Each episode consists of a support set allowing the model to adapt itself quickly to the unseen classes, and a query set on which the model is evaluated. In supervised FSL \cite{chen2019a, afrasiyabi2019associative, qiao2018few, ye2020few, dvornik2019diversity, li2020adversarial, sung2018learning, zhou2021binocular, singh2023transductive}, pretraining is done on a labeled dataset, whereas in unsupervised FSL \cite{psco,wang2022contrastive,unisiam, Qin2020ULDA, Antoniou2020Assume, khodadadeh2019unsupervised, hsu2018unsupervised, medina2020self, Shirekar_2023_WACV} the pretraining has to be conducted on an unlabeled dataset posing an extra challenge in the learning paradigm and neighboring these methods closer to the realm of self-supervised learning. Even though FSL is not of primal interest in this work, we assess the impact of \texttt{GeNIe} on a number of few-shot scenarios and state-of-the-art baselines by accentuating on its impact on the few-shot inference stage.

\section{Proposed Method: \texttt{GeNIe}
}
\label{sec:method}

%r \in (0,1)

Given a source image $X_S$ from category S = $\small{<} \texttt{source category}\small{>}$, we are interested in generating a target image $X_r$ from category $T = \small{<} \texttt{target category}\small{>}$. In doing so, we intend to ensure the low-level visual features or background context of the source image are preserved, so that we generate samples that would serve as \emph{hard negatives} for the \emph{source} image. To this aim, we adopt a conditional latent diffusion model (such as Stable Diffusion, \citep{rombach2021highresolution}) conditioned on a text prompt of the following format ``A photo of a $T = \small{<} \texttt{target category}\small{>}$''.
%, and $r$ is the ratio parameter that control the amount of noise and number iterations. Assuming that the standard diffusion model uses a noise standard deviation of $\sigma$ and $N$ for the total number of denoising steps, $r$ reduces them to $r\sigma$ and $\lfloor rN \rfloor$. Hence, $r=1$ and $r=0$ result in the standard diffusion model and an identity function, respectively. %Choosing a small $r$, results in a diffusion process that preserves some features of the source image and reduces the effect of the prompt.
%Let us assume we have a total of $N$ forward (and the same for backward) diffusion steps with $n \in (0, N]$. To associate these steps with equivalent noise levels, we define $\eta_n \in (0, 1] = n/N$. 

\textbf{Key Idea.} \genie{} in its basic form is a simple yet effective augmentation sample generator for improving a classifier $f_\theta(.)$ with the following two key aspects: (i) inspired by \citep{luzi2022boomerang,meng2021sdedit} instead of adding the full amount of noise $\sigma_{max}$ and going through all $N_{max}$ (being typically $50$) steps of denoising, we use less amount of noise ($r\sigma_{max}$, with $r \in (0,1)$) and consequently fewer number of denoising iterations ($\lfloor rN_{max}\rfloor$); (ii) we prompt the diffusion model with a $P$ mandating a target category $T$ different than the source $S$. Hence, we denote the conditional diffusion process as $X_r =$ \texttt{STDiff}$(X_S, P, r)$. In such a construct, the proximity of the final decoded image $X_r$ to the source image $X_S$ or the target category defined through the text prompt $P$ depends on $r$. Hence, by controlling the amount of noise, we can generate images that blend characteristics of both the text prompt $P$ and the source image $X_S$. If we do not provide much of visual details in the text prompt (e.g., desired background, etc.), we expect the decoded image $X_{r}$ to follow the details of $X_S$ while reflecting the semantics of the text prompt $P$. We argue, and demonstrate later, that the newly generated samples can serve as \emph{hard negative} examples for the source category $S$ since they share the low-level features of $X_S$ while representing the semantics of the target category, $T$. Notably, the source category $S$ can be randomly sampled or be carefully extracted from the confusion matrix of $f_\theta(.)$ based on real training data. The latter might result in even \emph{harder negative} samples being now cognizant of model confusions. Finally, we will append our initial dataset with the newly generated hard negative samples through \genie{} and (re)train the classifier model.

\begin{figure}[!t]
\centering
\includegraphics[width=1.0\linewidth]{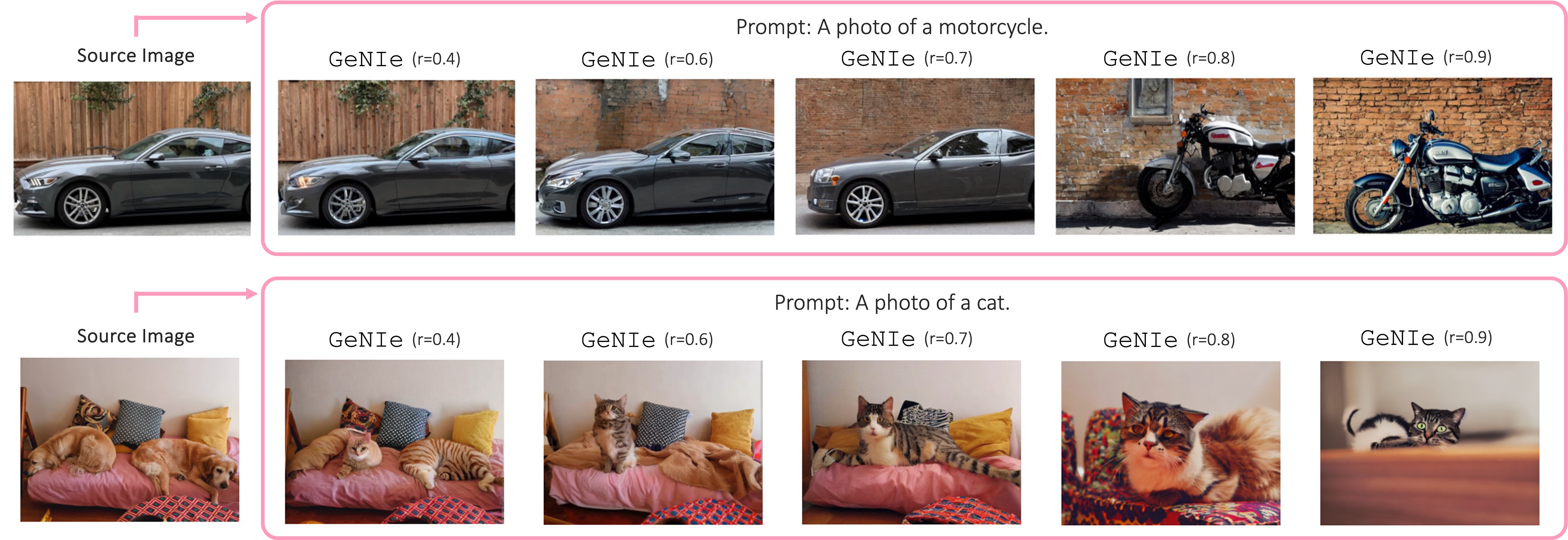}

\caption{\label{fig:noise_ab}{\bf Effect of noise ratio, $r$, in \genie{}:} we employ \genie{} to generate augmentations for the target classes (motorcycle and cat) with varying $r$. Smaller $r$ yields images closely resembling the source semantics, creating an inconsistency with the intended target label. By tracing $r$ from $0$ to $1$, augmentations gradually transition from source image characteristics to the target category. However, a distinct shift from the source to the target occurs at a specific $r$ that may vary for different source images or target categories. For more examples, please refer to Fig.~\ref{fig:noise_ab_supp}.}
% The generated images shift from car to motorcycle between $r \in (0.7, 0.8)$ in the top row and from dog to cat at $r \approx 0.4$ in the bottom row. }

\end{figure}

% \begin{wrapfigure}{D}{0.5\textwidth}
% \vspace{-0.5cm}
%     \begin{minipage}{0.6\textwidth}
%     \scalebox{0.86}{
%       \begin{algorithm}[H]
%     \caption{\scalebox{.9}{\texttt{GeNIe-Ada}}}\label{alg:genieada}
    
%     % \setstretch{.8}
%     \SetKwInOut{Input}{input}
%     \SetKwInOut{Output}{output}
%     \SetKwInput{Require}{Require}
% 	\SetKwInput{Return}{Return}
% 	\SetKw{Let}{let}
% 	\SetKwRepeat{Do}{do}{while}
	
% 	% \SetAlgoLined
% 	% \LinesNumbered
% 	\DontPrintSemicolon
% 	\SetNoFillComment
	
%     \Require{$X_S$, $X_T$, $f_\theta(.)$, \texttt{STDiff}$(.)$, $M$}
%     Extract $Z_S \gets f_\theta(X_s)$, $Z_T \gets f_\theta(X_T)$\;
    
%        \For{$ m \in [1, M]$} {
%             $r \gets \frac{m}{M}$, $Z_r \gets f_\theta(\,\texttt{STDiff}(X, P, r)\,)$\; 
%             $d_m \gets \frac{(Z_r-Z_S)^T(Z_T-Z_S)}{||Z_T-Z_S||_2}$
%             %Project embedding onto $\overrightarrow{ZZ_T}$:\;  
%             %$\bar{Z}_{\eta_m} \gets \texttt{Proj}(Z_{\eta_m}, \overrightarrow{ZZ_T})$
%     }
%     $m^{*} \gets \argmaxA_m |d_m-d_{m-1}|$, $\forall m \in [2, M]$\;
%     % \operatorname*{argmax}_m
%     $r^{*} \gets \frac{m^{*}}{n}$\;
%     %$\eta^{*} = \frac{\eta_{m^{*}} + \eta_{m^{*} - 1}}{2}$\;
%     \Return{$X_{r^{*}} = \texttt{STDiff}(X_S, P, r^{*})$}
% \end{algorithm}}

%     \end{minipage}
% \vspace{-0.3cm}
% \end{wrapfigure}
%%

\noindent\textbf{Enhancing \genie{}: \texttt{GeNIe-Ada}.} One of the remarkable aspects of \genie{} lies in its simple application, requiring only $X_S$, $P$, and $r$. However, selecting the appropriate value for $r$ poses a challenge as it profoundly influences the outcome. When $r$ is small, the resulting $X_r$ tends to closely resemble $X_S$, and conversely, when $r$ is large (closer to $1$), it tends to resemble the semantics of the target category. This phenomenon arises because a smaller noise level restricts the capacity of the diffusion model to deviate from the semantics of the input $X_S$. Thus, a critical question emerges: how can we select $r$ for a particular source image to generate samples that preserve the low-level semantics of the source category $S$ in $X_S$ while effectively representing the semantics of the target category $T$? We propose a method to determine an ideal value for $r$. 

Our intuition suggests that by varying the noise ratio $r$ from $0$ to $1$, $X_r$ will progressively resemble category $S$ in the beginning and category $T$ towards the end. However, somewhere between $0$ and $1$, $X_r$ will undergo a rapid transition from category $S$ to $T$. This phenomenon is empirically observed in our experiments with varying $r$, as depicted in Fig.~\ref{fig:noise_ab}. Although the exact reason for this rapid change remains uncertain, one possible explanation is that the intermediate points between two categories reside far from the natural image manifold, thus, challenging the diffusion model's capability to generate them. Ideally, we should select $r$ corresponding to just after this rapid semantic transition, as at this point, $X_r$ exhibits the highest similarity to the source image while belonging to the target category.

\begin{figure}[!t]
    \hspace{-0.9cm}
    \begin{minipage}{0.5\textwidth}
    % \vspace{-1.0cm}
        \centering
        \scalebox{.8}{
        \begin{algorithm}[H]
    \caption{\scalebox{.9}{\texttt{GeNIe-Ada}}}\label{alg:genieada}
    
    % \setstretch{.8}
    \SetKwInOut{Input}{input}
    \SetKwInOut{Output}{output}
    \SetKwInput{Require}{Require}
	\SetKwInput{Return}{Return}
	\SetKw{Let}{let}
	\SetKwRepeat{Do}{do}{while}
	
	% \SetAlgoLined
	% \LinesNumbered
	\DontPrintSemicolon
	\SetNoFillComment
	
    \Require{$X_S$, $X_T$, $f_\theta(.)$, \texttt{STDiff}$(.)$, $M$}
    Extract $Z_S \gets f_\theta(X_s)$, $Z_T \gets f_\theta(X_T)$\;
    
       \For{$ m \in [1, M]$} {
            $r \gets \frac{m}{M}$, $Z_r \gets f_\theta(\,\texttt{STDiff}(X, P, r)\,)$\; 
            $d_m \gets \frac{(Z_r-Z_S)^T(Z_T-Z_S)}{||Z_T-Z_S||_2}$
            %Project embedding onto $\overrightarrow{ZZ_T}$:\;  
            %$\bar{Z}_{\eta_m} \gets \texttt{Proj}(Z_{\eta_m}, \overrightarrow{ZZ_T})$
    }
    $m^{*} \gets \argmaxA_m |d_m-d_{m-1}|$, $\forall m \in [2, M]$\;
    % \operatorname*{argmax}_m
    $r^{*} \gets \frac{m^{*}}{n}$\;
    %$\eta^{*} = \frac{\eta_{m^{*}} + \eta_{m^{*} - 1}}{2}$\;
    \Return{$X_{r^{*}} = \texttt{STDiff}(X_S, P, r^{*})$}
\end{algorithm}}
    \end{minipage} \quad
    \hspace{-0.5cm}
    \begin{minipage}{.6\textwidth}
        \centering
        % \vspace{1.0cm}
        \includegraphics[width=0.88\linewidth]{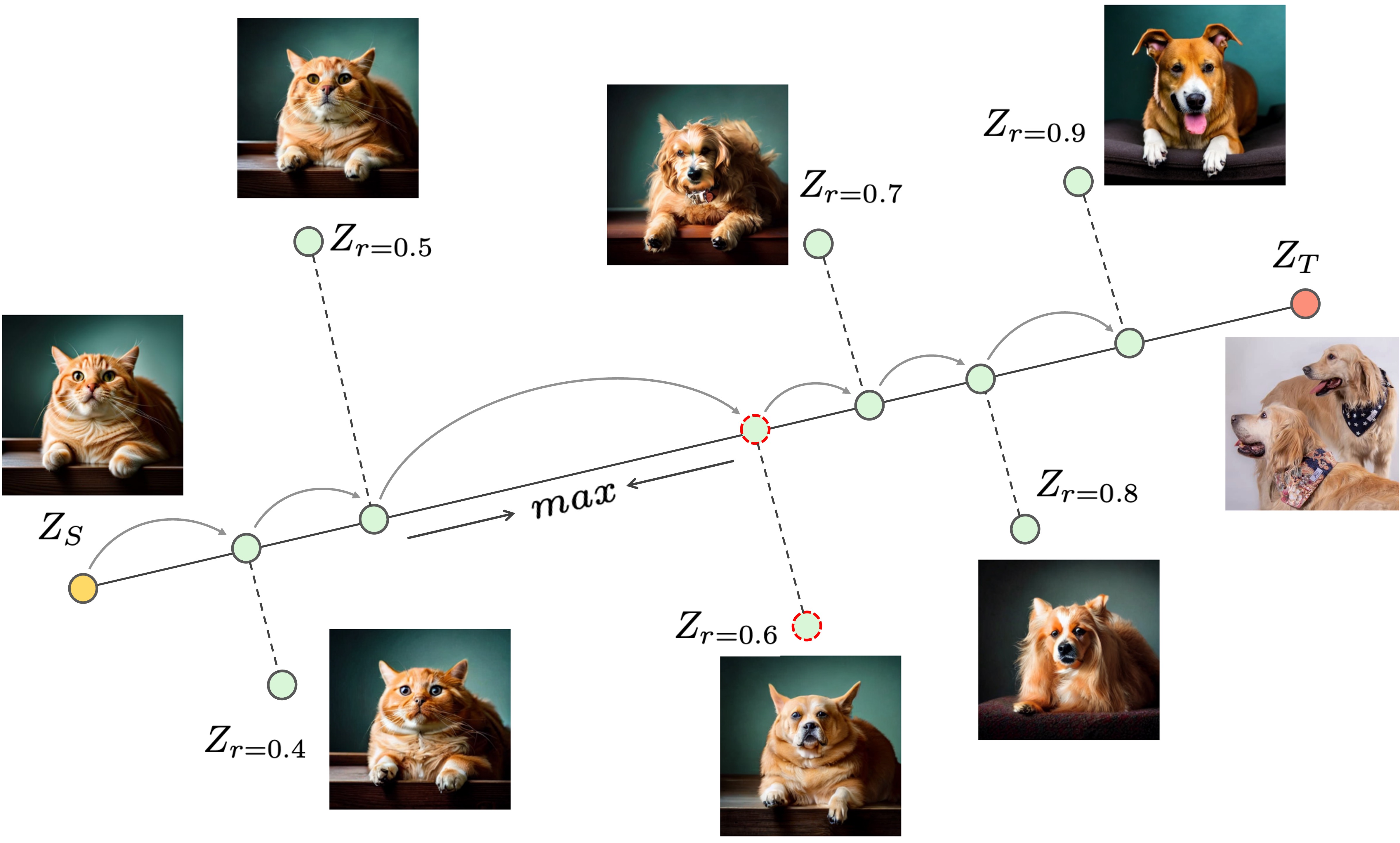}
        % \vspace{-.15in}
        % \caption{{\texttt{GeNIe-Ada} Noise Adaptive Selection}}

    \end{minipage}%
    \caption{\label{fig:geniepp}\texttt{GeNIe-Ada:} To choose $r$ adaptively for each (source image, target category) pair, we propose tracing the semantic trajectory from $Z_S$ (source image embeddings) to $Z_T$ (target embeddings) through the lens of the classifier $f_\theta(\cdot)$ (Algorithm \ref{alg:genieada}). We adaptively select the sample right after the largest semantic shift.}
    \vspace{-0.2in}
\end{figure}

We propose to trace the semantic trajectory between $X_S$ and $X_T$ through the lens of the classifier $f_\theta(.)$. As shown in Algorithm \ref{alg:genieada}, assuming access to the classifier backbone $f_{\theta}(.)$ and at least one example $X_T$ from the target category, we convert both $X_S$ and $X_T$ into their respective latent vectors $Z_S$ and $Z_T$ by passing them through $f_{\theta}(.)$. Then, we sample $M$ values for $r$ uniformly distributed $ \in (0,1)$, generating their corresponding $X_r$ and their latent vectors $Z_r$ for all those $r$. Subsequently, we calculate $d_r = \frac{(Z_r-Z_S)^T(Z_T-Z_S)}{||Z_T-Z_S||_2}$ as the distance between $Z_r$ and $Z_S$ projected onto the vector connecting $Z_S$ and $Z_T$.
%project the vector connecting $Z_S$ and $Z_r$ onto the vector connecting $X_S$ and $X_T$ to derive $\bar{Z_r}$, representing the projected distance of $X_r$ to $X_S$. Note that if we do not have access to any real image for $X_T$, we can still generate one using $\texttt{STDiff}$(X_S, P, r=1)$.
Our hypothesis posits that the rapid semantic transition corresponds to a sharp change in this projected distance. Therefore, we sample $n$ values for  $r$ uniformly distributed between $0$ and $1$, and analyze the variations in $d_r$. We identify the largest gap in $d_r$ and select the $r$ value just after the gap when increasing $r$, as detailed in Algorithm \ref{alg:genieada} and illustrated in Fig.~\ref{fig:geniepp}.

\vspace{-0.1in}
\section{Experiments}
\vspace{-0.1in}
Since the impact of augmentation is more pronounced when the training data is limited, we evaluate the impact of \genie{} on Few-Shot classification in Section~\ref{sec:few_shot}, Long-Tailed classification in Section~\ref{sec:long_tail}, and fine-grained classification in Section~\ref{sec:finegrained}.  For \genienoad{} in all scenarios, we utilize \genie{} to generate augmentations from the noise level set $\{0.5, 0.6, 0.7, 0.8, 0.9\}$. The selection of the appropriate noise level per source image and target is adaptive, achieved through Algorithm~\ref{alg:genieada}.

\textbf{Baselines.} We use Stable Diffusion 1.5 \citep{rombach2021highresolution} as our base diffusion model. In all settings, we use the same prompt format to generate images for the target class: i.e., ``A photo of a $\small{<}\texttt{target category}\small{>}$'', where we replace the $\texttt{target category}$ with the target category label. We generate $512\times512$ images for all methods. For fairness in comparison, we generate the same number of new images for each class. We use a single NVIDIA RTX $3090$ for image generation. We consider $4$ diffusion-based baselines and a suite of traditional data augmentation baselines.

\textbf{\imgimg} \citep{luzi2022boomerang,meng2021sdedit}: 
% This baseline follows the data augmentation strategy based on Stable Diffusion proposed in \citep{luzi2022boomerang,meng2021sdedit}. Concretely, 
We sample an image from a target class, add noise to its latent representation and then pass it along with a prompt for the target category through reverse diffusion. The focus here is on a target class for which we generate extra positive samples. Adding large amount of noise leads to generating an image less similar to the original image. We use two different noise magnitudes for this baseline: $r=0.3$ and $r=0.7$ and denote them by \posl{} and \posh{}, respectively.

\begin{figure}[!t]
\centering
\includegraphics[width=0.99\linewidth]{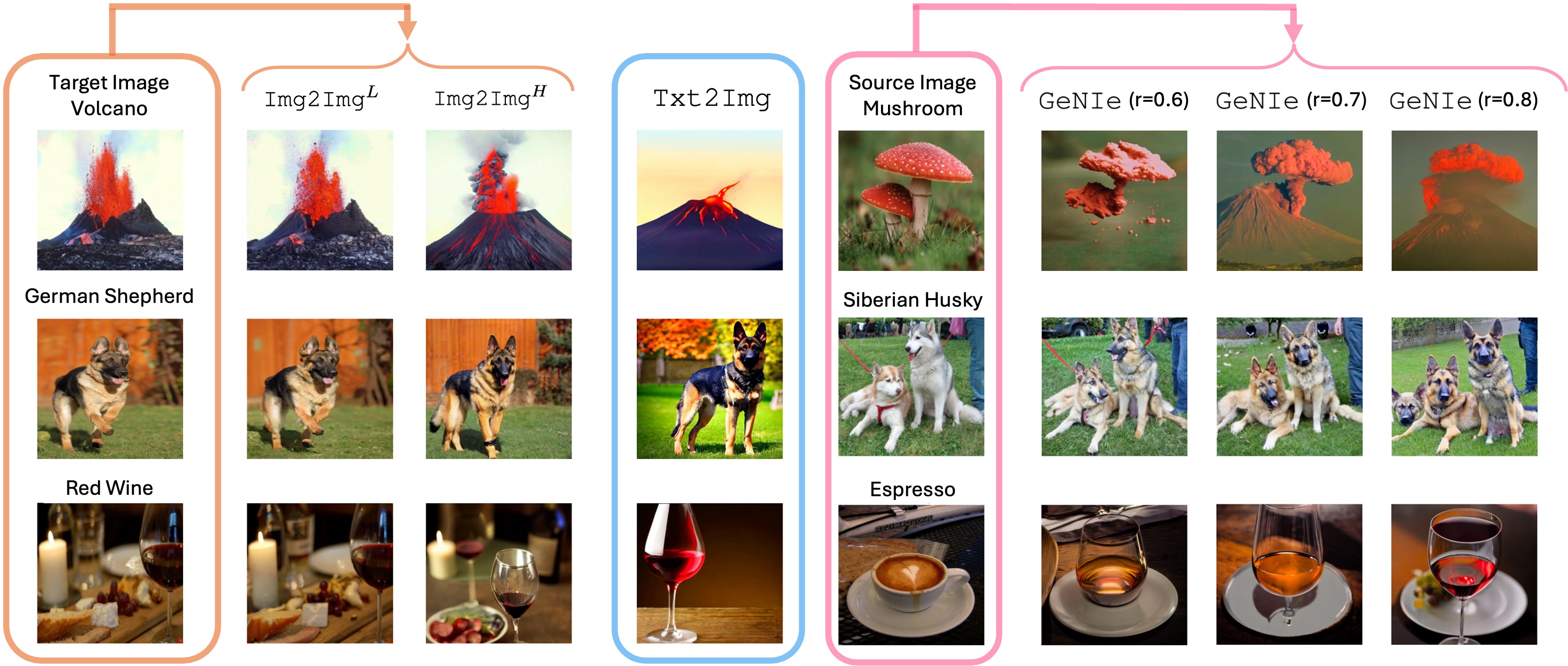}

\caption{{\label{fig:vis1_genie} \bf Visualization of Generative Samples:} We compare \genie{} with two baselines: \textbf{\posl{} augmentation:} both image and text prompt are from the same category. Adding noise does not change the image much, so they are not hard examples. \textbf{\txtimg{} augmentation:} We simply use the text prompt only to generate an image for the desired category (e.g., using a text2image method). Such images may be far from the domain of our task since the generation is not informed by any visual data from our task. \textbf{\genie{} augmentation:} We use the target category name in the text prompt only along with the source image. 
% We generate the desired images at appropriate amount of noise ($80\%$ in most cases).
%Note that lower amount of noise resulis still far from target category since the source image has to much of effect compared to the text prompt. 
% Note that we never define low-level features concretely and never evaluate if the augmented images preserves those features. That is our hypothesis only. We only evaluate if the augmented images help the accuracy.
}
\vspace{-0.2in}
\end{figure}

\textbf{\txtimg}~\citep{azizi2023synthetic,he2022synthetic}: For this baseline, we omit the forward diffusion process and only use the reverse process starting from a text prompt for the target class of interest. This is similar to the base text-to-image generation strategy adopted in \citep{rombach2021highresolution,he2022synthetic, shipard2023boosting, azizi2023synthetic, luo2023camdiff}. Fig.~\ref{fig:vis1_genie} 
% and \ref{fig:vis_genie_supp} 
illustrates a set of generated augmentation examples for \txtimg, \imgimg{}, and \genie.  
% Note that an extreme case of adding maximum noise ($r=0.999$) in \texttt{Img2Img} reduces it to \txtimg.

DAFusion \citep{trabucco2024effective}: In this method, an embedding is optimized with a set of images for each class to correspond to the classes in the dataset. This approach is introduced in Textual Inversion \citep{gal2022textual}. We optimize an embedding for 5000 iterations for each class in the dataset, followed by augmentation similar as the DAFusion method.

Cap2Aug\citep{roy2023cap2aug}: It is a recent diffusion-based data augmentation strategy that uses image captions as text prompts for an image-to-image diffusion model. 
% This method produces augmented images that are similar to the training images but offer semantic diversity.

\textbf{Traditional Data Augmentation:} We consider both weak and strong traditional augmentations. More specifically, for weak augmentation we use random resize crop with scaling $\in [0.2, 1.0]$ and horizontal flipping. For strong augmentation, we consider random color jitter, random grayscale, and Gaussian blur. For the sake of completeness, we also compare against data augmentations such as CutMix \citep{Cutmix} and MixUp \citep{mixup} that combine two images together. 

\subsection{Few-shot Classification}
\label{sec:few_shot}
We assess the impact of \genie{} compared to other augmentations in a number of few-shot classification (FSL) scenarios, where the model has to learn only from the samples contained in the ($N$-way, $K$-shot) support set and infer on the query set. Note that this corresponds to an inference-only FSL setting where a pretraining stage on an abundant dataset is discarded. The goal is to assess how well the model can benefit from the augmentations while keeping the original $N\times K$ samples intact. 

\textbf{Datasets.} We conduct our few-shot experiments on two most commonly adopted few-shot classification datasets: \textit{mini-}Imagenet \citep{Ravi2017OptimizationAA} and \textit{tiered-}Imagenet \citep{ren2018metalearning}. \textit{mini-}Imagenet is a subset of ImageNet \citep{deng2009imagenet} for few-shot classification. It contains $100$ classes with $600$ samples each. We follow the predominantly adopted settings of \citep{Ravi2017OptimizationAA, chen2019a} where we split the entire dataset into $64$ classes for training, $16$ for validation and $20$ for testing. \textit{tiered}-Imagenet is a larger subset of ImageNet with $608$ classes and a total of $779,165$ images, which are grouped into $34$ higher-level nodes in the \textit{ImageNet} human-curated hierarchy. This set of nodes is partitioned into $20$, $6$, and $8$ disjoint sets of training, validation, and testing nodes, and the corresponding classes form the respective meta-sets.
% \hamed{what sets we use?}

\textbf{Evaluation.} To quantify the impact of different augmentation methods, we evaluate the test-set accuracies of a state-of-the-art unsupervised few-shot learning method with \genie{} and compare them against the accuracies obtained using other augmentation methods. Specifically, we use UniSiam \citep{unisiam} pre-trained with ResNet-18, ResNet-34 and ResNet-50 backbones and follow its evaluation strategy of fine-tuning a logistic regressor to perform ($N$-way, $K$-shot) classification on the test sets of \textit{mini-} and \textit{tiered-}Imagenet. Following \citep{Ravi2017OptimizationAA}, an episode consists of a labeled support-set and an unlabelled query-set. The support-set contains $N$ randomly sampled classes where each class contains $K$ samples, whereas the query-set contains $Q$ randomly sampled unlabeled images per class. We conduct our experiments on the two most commonly adopted settings: ($5$-way, $1$-shot) and ($5$-way, $5$-shot) classification settings. Following the literature, we sample $16$-shots per class for the query set in both settings. We report the test accuracies along with the $95\%$ confidence interval over $600$ and $1000$ episodes for \textit{mini}-ImageNet and \textit{tiered}-ImageNet, respectively.

\begin{table*}[!t]
    % \caption{Global caption}
        \caption{\label{tab:mini_imagenet}{\bf \textit{mini}-ImageNet:} We use our augmentations on ($5$-way, $1$-shot) and ($5$-way, $5$-shot) few-shot settings of mini-Imagenet dataset with 3 different backbones (ResNet-18, 34, and 50). We compare with various baselines and show that our augmentations with UniSiam outperform all the baselines including \txtimg{} and DAFusion augmentation. The number of generated images per class is 4 for 1-shot and 20 for 5-shot settings. 
        % Note that UniSiam has used only weak augmentation, so we add the other methods for fair comparison.
        }
    \begin{minipage}[t]{.49\linewidth}
        %\caption{\textbf{Mini-ImageNet:}}
        
      \centering
        \scalebox{0.65}{
        \begin{tabular}{llccc}
    \toprule
    \multicolumn{5}{c}{\textbf{ResNet-18}} \\ \midrule
    \textbf{Augmentation} & \textbf{Method}  & \textbf{Pre-training} & \textbf{1-shot} & \textbf{5-shot}  \\
    \midrule
                       
   % - & $\Delta$-Encoder \citep{schwartz2018delta}  & sup. & 59.9 & 69.7 \\
    % - & SNCA \citep{wu2019improving} & sup.& 57.8$\pm$0.8 & 72.8$\pm$0.7 \\
    - & iDeMe-Net \cite{chen2019image} & sup. & 59.1$\pm$0.9 & 74.6$\pm$0.7 \\
    - & Robust + dist \cite{dvornik2019diversity} & sup. & 63.7$\pm$0.6 & 81.2$\pm$0.4 \\
    - & AFHN \cite{li2020adversarial} & sup. & 62.4$\pm$0.7 & 78.2$\pm$0.6 \\
    Weak & ProtoNet+SSL \cite{su2020when} & sup.+ssl & - & 76.6 \\
    Weak & Neg-Cosine \cite{liu2020negative} & sup. & 62.3$\pm$0.8 & 80.9$\pm$0.6 \\
    - & Centroid Align\cite{afrasiyabi2019associative}  & sup. & 59.9$\pm$0.7 & 80.4$\pm$0.7 \\
    - & Baseline \cite{chen2019a} & sup. &  59.6$\pm$0.8  & 77.3$\pm$0.6 \\
    - & Baseline++ \cite{chen2019a} & sup. &  59.0$\pm$0.8 & 76.7$\pm$0.6 \\
    Weak & PSST \cite{chen2021pareto} & sup.+ssl & 59.5$\pm$0.5 & 77.4$\pm$0.5 \\
    % \cmidrule(r){2-6}
    \midrule
    Weak & UMTRA \cite{khodadadeh2019unsupervised} & unsup.  & 43.1$\pm$0.4 & 53.4$\pm$0.3 \\
    Weak & ProtoCLR  \cite{medina2020self} & unsup.  & 50.9$\pm$0.4 & 71.6$\pm$0.3 \\
    Weak & SimCLR \cite{chen2020a} & unsup.  &  62.6$\pm$0.4 & 79.7$\pm$0.3 \\
    Weak & SimSiam \cite{chen2020exploring} & unsup.  & 62.8$\pm$0.4 & 79.9$\pm$0.3\\
   % \cmidrule(r){2-6}
    Weak & UniSiam+dist \cite{unisiam} & unsup.  & \textbf{64.1$\pm$0.4}   & \textbf{82.3$\pm$0.3}  \\
    \midrule
    Weak & UniSiam \cite{unisiam} & unsup.  & 63.1$\pm$0.8 & 81.4$\pm$0.5 \\
    % \cmidrule(r){2-6}                 
    Strong & UniSiam \cite{unisiam} & unsup.  & 62.8$\pm$0.8 & 81.2$\pm$0.6  \\
    CutMix \cite{Cutmix} & UniSiam \cite{unisiam}&  unsup.  & 62.7$\pm$0.8	& 80.6$\pm$0.6  \\
    MixUp \cite{mixup} & UniSiam \cite{unisiam} & unsup.  & 62.1$\pm$0.8 & 80.7$\pm$0.6  \\
    \posl \cite{luzi2022boomerang} & UniSiam \cite{unisiam} & unsup.  & 63.9$\pm$0.8 & 82.1$\pm$0.5  \\
    \posh \cite{luzi2022boomerang} & UniSiam \cite{unisiam} & unsup.  & 69.1$\pm$0.7 & 84.0$\pm$0.5  \\
    \txtimg \cite{azizi2023synthetic,he2022synthetic} & UniSiam \cite{unisiam} & unsup.  & 74.1$\pm$0.6 & 84.6$\pm$0.5  \\
    DAFusion \cite{trabucco2024effective} & UniSiam \cite{unisiam} & unsup.  & 64.3$\pm$1.8 & 82.0$\pm$1.4  \\
    \rowcolor{LightCyan}
    \genie{} (Ours) & UniSiam \cite{unisiam}  & unsup.  & \textbf{75.5$\pm$0.6} & \textbf{85.4$\pm$0.4}  \\
    \rowcolor{LightCyan}
    \genienoad{} (Ours) & UniSiam \cite{unisiam} & unsup.  & \textbf{76.8$\pm$0.6} & \textbf{85.9$\pm$0.4}  \\
    \bottomrule
    
  \end{tabular}
        }
    \end{minipage} \quad
    \begin{minipage}[t]{.50\linewidth}
        %\caption{\textbf{} }
        
        \centering
        \scalebox{0.55}{
        \begin{tabular}{llccc}
    \toprule
    \multicolumn{5}{c}{\textbf{ResNet-34}} \\ \midrule
    \textbf{Augmentation} & \textbf{Method} & \textbf{Pre-training} & \textbf{1-shot} & \textbf{5-shot}  \\
    \midrule
                       
    % Weak & MatchingNet \cite{vinyals2016matching} & sup. &  53.2$\pm$0.8  & 68.3$\pm$0.7 \\
    % Weak & ProtoNet \cite{snell2017prototypical}  & sup. &  53.9$\pm$0.8  & 74.7$\pm$0.6 \\
    % Weak & MAML \cite{finn2017model} & sup. & 51.5$\pm$0.9  & 65.9$\pm$0.8 \\
    % Weak & RelationNet \cite{sung2018learning}  & sup. &  51.7$\pm$0.8  & 69.6$\pm$0.7 \\
    Weak & Baseline \cite{chen2019a} & sup. &  49.8$\pm$0.7  & 73.5$\pm$0.7 \\
    Weak & Baseline++ \cite{chen2019a} & sup. &  52.7$\pm$0.8  & 76.2$\pm$0.6 \\
    \midrule
    Weak & SimCLR \cite{chen2020a} & unsup. & 64.0$\pm$0.4 & 79.8$\pm$0.3 \\
    Weak & SimSiam \cite{chen2020exploring}  & unsup.  & 63.8$\pm$0.4 &80.4$\pm$0.3 \\
     % \cmidrule(r){2-6}
    Weak & UniSiam+dist \cite{unisiam} & unsup. & \textbf{65.6$\pm$0.4} & \textbf{83.4$\pm$0.2} \\
    \midrule
    Weak & UniSiam \cite{unisiam} & unsup. & 64.3$\pm$0.8 & 82.3$\pm$0.5 \\
    % \cmidrule(r){2-6}
    Strong & UniSiam \cite{unisiam} & unsup.  & 64.5$\pm$0.8 & 82.1$\pm$0.6  \\
    CutMix \cite{Cutmix} & UniSiam \cite{unisiam} & unsup.  & 64.0$\pm$0.8	& 81.7$\pm$0.6  \\
    MixUp \cite{mixup} & UniSiam \cite{unisiam} & unsup.  & 63.7$\pm$0.8 & 80.1$\pm$0.8  \\
    \posl \cite{luzi2022boomerang} & UniSiam \cite{unisiam} & unsup.  & 65.5$\pm$0.8 & 82.9$\pm$0.5  \\
    \posh \cite{luzi2022boomerang} & UniSiam \cite{unisiam} & unsup.  & 70.5$\pm$0.8 & 84.8$\pm$0.5  \\
    \txtimg \cite{azizi2023synthetic,he2022synthetic} & UniSiam \cite{unisiam} & unsup.  & 75.4$\pm$0.6 & 85.5$\pm$0.5  \\
    DAFusion \cite{trabucco2024effective} & UniSiam \cite{unisiam} & unsup.  & 64.7$\pm$1.9 & 83.2$\pm$1.4  \\
    \rowcolor{LightCyan}
    \genie{} (Ours) & UniSiam \cite{unisiam} & unsup.  & \textbf{77.1$\pm$0.6} & \textbf{86.3$\pm$0.4}  \\
    \rowcolor{LightCyan}
    \genienoad{} (Ours) & UniSiam \cite{unisiam} & unsup.  & \textbf{78.5$\pm$0.6} & \textbf{86.6$\pm$0.4}  \\
    \toprule
    \multicolumn{5}{c}{\textbf{ResNet-50}} \\ \midrule
    Weak & PDA+Net \cite{chen2021few} & unsup. & 63.8$\pm$0.9 & 83.1$\pm$0.6 \\
    Weak & Meta-DM \cite{hu2023metadm}& unsup. & 66.7$\pm$0.4 & 85.3$\pm$0.2 \\
    \midrule
    Weak & UniSiam \cite{unisiam} & unsup. & 64.6$\pm$0.8 & 83.4$\pm$0.5 \\
    Strong & UniSiam \cite{unisiam} & unsup.  & 64.8$\pm$0.8 & 83.2$\pm$0.5  \\

    CutMix \cite{Cutmix} & UniSiam \cite{unisiam} & unsup.  & 64.3$\pm$0.8 & 83.2$\pm$0.5  \\
    MixUp \cite{mixup} & UniSiam \cite{unisiam} & unsup.  & 63.8$\pm$0.8 & 84.6$\pm$0.5  \\
    \posl \cite{luzi2022boomerang} & UniSiam \cite{unisiam} & unsup.  & 66.0$\pm$0.8 & 84.0$\pm$0.5  \\
    \posh \cite{luzi2022boomerang} & UniSiam \cite{unisiam} & unsup.  & 71.1$\pm$0.7 & 85.7$\pm$0.5  \\
    \txtimg \cite{azizi2023synthetic,he2022synthetic} & UniSiam \cite{unisiam} & unsup.  & 76.4$\pm$0.6 & 86.5$\pm$0.4  \\
    DAFusion \cite{trabucco2024effective} & UniSiam \cite{unisiam} & unsup.  & 65.7$\pm$1.8 & 83.9$\pm$1.2  \\
    \rowcolor{LightCyan}
    \genie{} (Ours) & UniSiam \cite{unisiam} & unsup.  & \textbf{77.3$\pm$0.6} & \textbf{87.2$\pm$0.4}  \\
    \rowcolor{LightCyan}
    \genienoad{} (Ours) & UniSiam \cite{unisiam}& unsup.  & \textbf{78.6$\pm$0.6} & \textbf{87.9$\pm$0.4}  \\
    \bottomrule
  \end{tabular}
        }
    \end{minipage}%
\vspace{-0.2in}
\end{table*}

\textbf{Implementation Details:} \genie{} generates augmented images for each class using images from all other classes as the source image. We use $r=0.8$ in our experiments. % along with their class labels as the corresponding text prompt. 
We generate $4$ samples per class as augmentations in the $5$-way, $1$-shot setting and $20$ samples per class as augmentations in the $5$-way, $5$-shot setting. For the sake of a fair comparison, we ensure that the total number of labelled samples in the support set after augmentation remains the same across all different traditional and generative augmentation methodologies. Due to the expensive training of embeddings for each class in each episode, we only evaluated the DA-Fusion baseline on the first 100 episodes.

\textbf{Results:} The results on \textit{mini-}Imagenet and \textit{tiered-}Imagenet for both ($5$-way, $1$ and $5$-shot) settings are summarized in Table~\ref{tab:mini_imagenet} and Table~\ref{tab:sota_tiered}, respectively. Regardless of the choice of backbone, we observe that \genie{} helps consistently improve UniSiam's performance and outperform other supervised and unsupervised few-shot classification methods as well as other diffusion-based \citep{trabucco2024effective, luzi2022boomerang, stable, he2022synthetic} and classical \citep{Cutmix, mixup} data augmentation techniques on both datasets, across both ($5$-way, $1$ and $5$-shot) settings. Our noise adaptive method of selecting optimal augmentations per source image (\genienoad{}) further improves \genie's performance across all three backbones, both few-shot settings, and both datasets (\textit{mini} and \textit{tiered-}Imagenet). Few-shot accuracies for ResNet-34 computed on \textit{tiered}Imagenet are reported in Section \ref{sec:r34tiered} of the appendix. Note that employing CutMix and MixUp seems to lead to performance degradation compared to weak augmentations, probably due to overfitting since these methods can only choose from $4$ other classes to mix.
%While these techniques introduce noise beneficial for regularizing models on large datasets, their impact can be counterproductive in few-shot learning, particularly due to the limited number of training examples. In scenarios with constrained labeled data, where semantic content is paramount, the blending introduced by CutMix and MixUp may result in the loss of crucial information, posing challenges for models in discerning important patterns. Hence, the importance of tailored data augmentation strategies is underscored in the context of few-shot learning.

\subsection{Long-Tailed Classification}
\label{sec:long_tail}

We evaluate our method on long-tailed data, where the number of instances per class is unbalanced, with most categories having limited samples (tail). Our goal is to mitigate this bias by augmenting the tail of the distribution with generated samples. We evaluate \genie{} using two different backbones and methods: the ViT architecture with LViT \citep{xu2023learning}, and ResNet50 with VL-LTR \citep{tian2022vl}.

Following LViT \citep{xu2023learning}, we first train an MAE \citep{he2021masked} and ViT on the unbalanced dataset without any augmentation. Next, we train the Balanced Fine-Tuning stage of LViT by incorporating the augmentation data generated using \genie{} or other baselines. For ResNet50, we use VL-LTR code to fine-tune the CLIP \citep{radford2021learning} ResNet50 pretrained backbone with generated augmentations by \genie{}.

% We following the proposed settings in \citep{xu2023learning} for the Balanced Fine Tuning stage, which includes traditional augmentation including CutMix, MixUp with a Balanced Binary Cross-Entropy (Bal-BCE) loss. \hamed{this sentence is not finished,}

\textbf{Dataset:} We perform experiments on ImageNet-LT \citep{OLTR}. It contains $115.8$K images from $1,000$ categories. The number of images per class varies from $1280$ to $5$. %Max and Min number of images per class is $1280$ and $5$ respectively. 
Imagenet-LT classes can be divided into $3$ groups: ``Few'' with less than $20$ images, ``Med'' with $20-100$ images, and ``Many'' with more than $100$ images. Imagenet-LT uses the same validation set as ImageNet. We augment ``Few'' categories only and limit the number of generated images to $50$ samples per class. For \genie{}, instead of randomly sampling the source images from other classes, we use a confusion matrix on the training data to find the top-$4$ most confused classes and only consider those classes for random sampling of the source image. The source category may be from ``Many'', ``Med'', or ``Few sets''.

% \textbf{Implementation Details:} We download the pretrained ViT-B of LViT \citep{xu2023learning} and finetune it with Bal-BCE loss proposed therein on the augmented dataset. Training takes 2 hours on four NVIDIA RTX 3090 GPUs. We use the same hyperparameters as in \citep{xu2023learning} for finetuning: $100$ epochs, $lr=0.008$, batch size of $1024$, CutMix and MixUp for the data augmentation.   

\textbf{Results:} Augmenting training data with \genienoad{} improves accuracy on the ``Few'' set by $11.7\%$ and $4.4\%$ compared with LViT only and LViT with \txtimg{} augmentation baselines respectively. 
% \genienoad{} improves the overall accuracy by $2.2\%$ compared to the baselines with only traditional augmentations.
In ResNet50, \genienoad{} outperforms Cap2Aug baseline in ``Few'' categories by $7.6\%$. The results are summarized in Table~\ref{tab:imagenet_lt}. Please refer to Section ~\ref{sec:appendix_longtail} for implementation details.

% \begin{table}[t]
% \centering
% \caption{\label{tab:finegrained} \textbf{Few-shot Learning on Finegrained dataset:} We train SVM classifier on top of DinoV2 ViT-G pretrained backbone and report Top-1 accuracy on the test set of each dataset. Our baseline here is the SVM trained with traditional augmentations. \genie{} outperforms Baseline across all dataset. Compared to data augmentation with \txtimg{} only, \genie{} improves accuracy by $6.6$ point and $5.2$ point in Cars and Aircraft dataset respectively. }
% \scalebox{0.9}{
% \begin{tabular}{lcccc}
% \toprule 

% \multicolumn{1}{l}{Method} & Birds  & Cars & Foods & Aircraft \\

% \multicolumn{1}{l}{} &  \small{CUB200 \citep{Wah11thecaltech-ucsd}} & \small{Cars196 \citep{carsdataset}}  & \small{Food101 \citep{food101}} &  \small{Aircraft \citep{aircraft}} \\ \midrule
% Baseline & 90.34 & 49.77 & 82.92 & 29.19 \\
% \posl  & 90.73 & 50.38 & 87.4 & 30.99 \\ 
% \posh  & 91.31 & 56.39 & 91.7 & 34.73 \\ 
% \txtimg  & 92.01 & 81.29 & 92.95 & 41.73 \\ 
% \rowcolor{LightCyan}
% \genie{} (r=$0.8$) & 92.49 & 87.74 & \textbf{93.06} & 46.49 \\ 
% \rowcolor{LightCyan}
% \genie{} (r=$0.7$) & \textbf{92.52} & \textbf{87.92} & 92.88 & \textbf{47.01} \\ 
% \bottomrule
% \end{tabular}}

% \end{table}

\begin{table*}[!t]
    % \caption{Global caption}
    \begin{minipage}[t]{.5\linewidth}
        \caption{\label{tab:sota_tiered} \textbf{\textit{tiered-}ImageNet:} Accuracies (\% $\pm$ std) for $5$-way, $1$-shot and $5$-way, $5$-shot classification settings on the test-set. We compare against various SOTA supervised and unsupervised few-shot classification baselines as well as other augmentation methods, with UniSiam \cite{unisiam} pre-trained ResNet-18,50 backbones.}
        
      \centering
        \scalebox{0.57}{
        \begin{tabular}{llccc}
    \toprule
    \multicolumn{5}{c}{\textbf{ResNet-18}} \\ \midrule
   \textbf{Augmentation} & \textbf{Method}   & \textbf{Pre-training} &  \textbf{1-shot}     & \textbf{5-shot}  \\
    \midrule

     % - & Transd-CNAPS \cite{Bateni2022_TransductiveCNAPS}   & sup. & 65.9$\pm$1.0 & 81.8$\pm$0.7 \\
     % - & FEAT \cite{ye2020fewshot}    & sup. & 70.8 & 84.8 \\
     Weak & SimCLR\cite{chen2020a}     & unsup. & 63.4$\pm$0.4  & 79.2$\pm$0.3\\
     Weak & SimSiam \cite{chen2020exploring}     & unsup.     &    64.1$\pm$0.4    &  81.4$\pm$0.3 \\
     \midrule
     % Weak & UniSiam + dist \cite{unisiam}  & unsup.  & 67.0$\pm$0.4 & 84.5$\pm$0.3 \\
    
    Weak & UniSiam \cite{unisiam}  & unsup.  & 63.1$\pm$0.7 & 81.0$\pm$0.5 \\
    Strong & UniSiam \cite{unisiam}    & unsup.  & 62.8$\pm$0.7 & 80.9$\pm$0.5 \\
    CutMix \cite{Cutmix} & UniSiam \cite{unisiam}    & unsup.  & 62.1$\pm$0.7 & 78.9$\pm$0.6 \\
    MixUp \cite{mixup} & UniSiam \cite{unisiam}  & unsup.  & 62.1$\pm$0.7 & 78.4$\pm$0.6 \\
    \posl \cite{luzi2022boomerang} & UniSiam \cite{unisiam}   & unsup.  & 63.9$\pm$0.7 & 81.8$\pm$0.5 \\
    \posh \cite{luzi2022boomerang} & UniSiam \cite{unisiam}  & unsup.  & 68.7$\pm$0.7 & 83.5$\pm$0.5 \\
    \txtimg \cite{he2022synthetic} & UniSiam \cite{unisiam}  & unsup.  & 72.9$\pm$0.6 & 84.2$\pm$0.5 \\
    DAFusion \cite{trabucco2024effective} & UniSiam \cite{unisiam} & unsup.  & 62.6$\pm$2.1 & 81.0$\pm$1.5  \\
    \rowcolor{LightCyan}
    \genie (Ours) & UniSiam \cite{unisiam}  & unsup.  & \textbf{73.6$\pm$0.6} & \textbf{85.0$\pm$0.4} \\
    \rowcolor{LightCyan}
     \genienoad (Ours) & UniSiam \cite{unisiam}  & unsup.  & \textbf{75.1$\pm$0.6} & \textbf{85.5$\pm$0.5} \\

   %  \midrule
    % \multicolumn{5}{c}{\textbf{ResNet-34 }} \\ \midrule
    % Weak & UniSiam + dist \cite{unisiam}   & unsup. & 68.7$\pm$0.4 & 85.7$\pm$0.3 \\
    % Weak & UniSiam \cite{unisiam}   & unsup. & 65.0$\pm$0.7 & 82.5$\pm$0.5 \\
    % Strong & UniSiam \cite{unisiam}  & unsup.  & 64.8$\pm$0.7 & 82.4$\pm$0.5 \\
    % CutMix \cite{Cutmix} & UniSiam \cite{unisiam}  & unsup.  & 63.8$\pm$0.7 & 80.3$\pm$0.6 \\
    % MixUp \cite{mixup} & UniSiam \cite{unisiam} & unsup.  & 64.1$\pm$0.7 & 80.0$\pm$0.6 \\
    % \posl \cite{luzi2022boomerang} & UniSiam \cite{unisiam}  & unsup.  & 66.1$\pm$0.7 & 83.1$\pm$0.5 \\
    % \posh \cite{luzi2022boomerang} & UniSiam \cite{unisiam} & unsup.  & 70.4$\pm$0.7 & 84.7$\pm$0.5 \\
    % \txtimg \cite{he2022synthetic} & UniSiam \cite{unisiam} & unsup.  & 75.0$\pm$0.6 & 85.4$\pm$0.4 \\
    % DAFusion \cite{trabucco2024effective} & UniSiam \cite{unisiam} & unsup.  & 64.1$\pm$2.1 & -$\pm$-  \\
    % \rowcolor{LightCyan}
    % \genie{} (Ours) & UniSiam \cite{unisiam}  & unsup.  & \textbf{75.7$\pm$0.6} & \textbf{86.0$\pm$0.4} \\
    %  \rowcolor{LightCyan}
    % \genienoad{} (Ours) & UniSiam \cite{unisiam}  & unsup.  & \textbf{76.9$\pm$0.6} & \textbf{86.3$\pm$0.2} \\
    \midrule    
    \multicolumn{5}{c}{\textbf{ResNet-50 }} \\ \midrule

    Weak & PDA+Net \cite{chen2021few} & unsup. & 69.0$\pm$0.9 & 84.2$\pm$0.7 \\
    Weak & Meta-DM \cite{hu2023metadm} & unsup. & 69.6$\pm$0.4 & 86.5$\pm$0.3 \\
    \midrule
    Weak & UniSiam + dist \cite{unisiam}  & unsup. & 69.6$\pm$0.4 & 86.5$\pm$0.3 \\
    Weak & UniSiam \cite{unisiam}  & unsup.     & 66.8$\pm$0.7 & 84.7$\pm$0.5  \\
    Strong & UniSiam \cite{unisiam} & unsup.  & 66.5$\pm$0.7 & 84.5$\pm$0.5 \\
    CutMix \cite{Cutmix} & UniSiam \cite{unisiam}  & unsup.  & 66.0$\pm$0.7 & 83.3$\pm$0.5 \\
    MixUp \cite{mixup} &UniSiam \cite{unisiam} & unsup.  & 66.1$\pm$0.5 & 84.1$\pm$0.8 \\
    \posl \cite{luzi2022boomerang} &UniSiam \cite{unisiam}  & unsup.  & 67.8$\pm$0.7 & 85.3$\pm$0.5 \\
    \posh \cite{luzi2022boomerang} &UniSiam \cite{unisiam}  & unsup.  & 72.4$\pm$0.7 & 86.7$\pm$0.4 \\
    \txtimg \cite{he2022synthetic} & UniSiam \cite{unisiam} & unsup.  & 77.1$\pm$0.6 & 87.3$\pm$0.4 \\
    DAFusion \cite{trabucco2024effective} & UniSiam \cite{unisiam} & unsup.  & 66.5$\pm$2.2 & 84.8$\pm$1.4  \\
    \rowcolor{LightCyan}
    \genie{} (Ours)  & UniSiam \cite{unisiam}   & unsup.  & \textbf{78.0$\pm$0.6} & \textbf{88.0$\pm$0.4} \\
    \rowcolor{LightCyan}
    \genienoad{} (Ours) & UniSiam \cite{unisiam}    & unsup.  & \textbf{78.8$\pm$0.6} & \textbf{88.6$\pm$0.6} \\
    
    \bottomrule
  \end{tabular}
        
        }
    \end{minipage} \quad
    \begin{minipage}[t]{.45\linewidth}
        \caption{\label{tab:imagenet_lt} \textbf{Long-Tailed ImageNet-LT:} We compare different augmentation methods on ImageNet-LT and report Top-1 accuracy for ``Few'', ``Medium'', and ``Many'' sets.  
        % $*$: indicates training with 384 resolution so is not directly comparable with other methods with 224 resolution.
        On the ``Few'' set and LiVT method, our augmentations improve the accuracy by $11.7$ points compared to LiVT original augmentation and $4.4$ points compared to \txtimg{}. \genienoad{} outperforms Cap2Aug baseline in ``Few'' categories by $7.6\%$. Refer to Table~\ref{tab:imagenet_lt_apendix} for a full comparison with prior Long-Tailed methods. }
        
        \centering
        \scalebox{0.65}{

        \begin{tabular}{lccc|c}
\toprule
\multicolumn{5}{c}{\textbf{ResNet-50}} \\\midrule
\multicolumn{1}{l}{Method}  & Many & Med.        & Few           & Overall Acc  \\ \midrule
% CE \cite{CB} & 224 & 64.0 & 33.8   & 5.8  & 41.6 \\
% LDAM \cite{LDAM} & 224& 60.4 & 46.9   & 30.7 & 49.8 \\
% c-RT \cite{NCM}  & 224 & 61.8 & 46.2   & 27.3 & 49.6 \\
% $\tau$-Norm \cite{NCM}  & 224 & 59.1 & 46.9   & 30.7 & 49.4 \\
% Causal \cite{CausalNorm}  & 224& 62.7 & 48.8   & 31.6 & 51.8 \\
% Logit Adj. \cite{LA} & 224& 61.1 & 47.5   & 27.6 & 50.1 \\
% RIDE(4E)$\dagger$ \cite{RIDE}  & 224& 68.3 & 53.5   & 35.9 & 56.8 \\
% MiSLAS \cite{MiSLAS}  & 224& 62.9	& 50.7	 & 34.3	& 52.7 \\
% DisAlign \cite{DisAlign}  & 224& 61.3 & 52.2   & 31.4 & 52.9 \\
% ACE$\dagger$ \cite{ACE} & 224& 71.7 & 54.6   & 23.5 & 56.6 \\
% PaCo$\dagger$ \cite{PaCo}  & 224& 68.0 & 56.4   & 37.2 & 58.2 \\
% TADE$\dagger$ \cite{TADE}  & 224& 66.5 & \textbf{57.0}   & 43.5 & 58.8 \\
% TSC \cite{TSC}  & 224& 63.5 & 49.7   & 30.4 & 52.4 \\
% GCL \cite{GCL}  & 224& 63.0 & 52.7   & 37.1 & 54.5 \\
% TLC \cite{TLC}  & 224& 68.9 & 55.7   & 40.8 & 55.1 \\
% BCL$\dagger$ \cite{BCL}  & 224& 67.6 & 54.6   & 36.6 & 57.2 \\
% NCL \cite{NCL}  & 224& 67.3 & 55.4   & 39.0 & 57.7 \\
% SAFA \cite{SAFA}  & 224& 63.8 & 49.9   & 33.4 & 53.1 \\
% DOC \cite{DOC} & 224& 65.1 & 52.8   & 34.2 & 55.0 \\
% DLSA \cite{DLSA}  & 224 & 67.8 & 54.5   & 38.8 & 57.5 \\ \midrule

{ResLT~\cite{cui2022reslt}}  & {63.3} & {53.3} & {40.3} & {55.1}\\
  {PaCo~\cite{cui2021parametric}}  & {68.2} & {58.7} & {41.0} & {60.0}\\
  {LWS~\cite{kang2019decoupling}}  & {62.2} & {48.6} & {31.8} & {51.5}\\
  {Zero-shot CLIP~\cite{radford2021learning}}  & {60.8} & {59.3} & {58.6} & {59.8}\\ 
 {DRO-LT~\cite{samuel2021distributional} } & {64.0} & {49.8} & {33.1} & {53.5}\\
 {VL-LTR~\cite{tian2022vl}}  & {77.8} & {67.0} & {50.8} & {70.1}\\
 Cap2Aug \cite{roy2023cap2aug}  & {78.5} & {\textbf{67.7}} & {51.9} & {70.9}\\  	
 \rowcolor{LightCyan}
 \genienoad{}  & {\textbf{79.2}} & {64.6} & {\textbf{59.5}} & {\textbf{71.5}}\\
 
 \toprule
\multicolumn{5}{c}{\textbf{ViT-B}} \\\midrule
\multicolumn{1}{l}{Method}  & Many & Med.        & Few           & Overall Acc  \\ 
% \midrule
% LiVT \cite{xu2023learning}& 384 & 76.4 & 59.7  & 42.7 & 63.8 \\
\midrule
ViT \cite{ViT}  &  50.5 & 23.5   & 6.9 & 31.6 \\
MAE \cite{MAE}  &  74.7 & 48.2   & 19.4 & 54.5 \\
DeiT \cite{deit3}  &  70.4     & 40.9     & 12.8     & 48.4 \\
LiVT \cite{xu2023learning}  &  73.6 & 56.4   & 41.0 & 60.9 \\
LiVT + \posl   & 74.3 & 56.4 & 34.3 & 60.5\\
LiVT + \posh  &  73.8 & 56.4 & 45.3 & 61.6\\
LiVT + \txtimg   &  \textbf{74.9} & 55.6 & 48.3 & 62.2\\
\rowcolor{LightCyan}
% LiVT + \genie{} (r=$0.8$) & 224 &  74.5 & 56.7 & 50.9 & 62.8\\
\rowcolor{LightCyan}
LiVT + \genienoad{} &  74.0 & \textbf{56.9} & \textbf{52.7} & \textbf{63.1}\\
\bottomrule
\end{tabular}
  
        }

%         \vspace{0.05cm}
%         \caption{\label{tab:finegrained} \textbf{Few-shot Learning on Fine-grained dataset:} We utilize a SVM classifier trained atop the DinoV2 ViT-G pretrained backbone, reporting Top-1 accuracy for the test set of each dataset. The baseline is an SVM trained on the same backbone using weak augmentation. Across all datasets, \genie{} surpasses this baseline.}
%         % \vspace{-0.25cm}
%         % Compared to data augmentation with \txtimg{} only, \genie{} improves accuracy by $6.6$ point and $5.2$ point in Cars and Aircraft dataset respectively. }
% \scalebox{0.63}{
% \begin{tabular}{lcccc}
% \toprule 

% % \multicolumn{1}{l}{Method} & Birds   & Cars & Foods & Aircraft \\

% \multicolumn{1}{l}{Method} &  \small{CUB200 \cite{Wah11thecaltech-ucsd}} & \small{Cars196 \cite{carsdataset}}  & \small{Food101 \cite{food101}} &  \small{Aircraft \cite{aircraft}} \\ \midrule
% Baseline & 90.3 & 49.8 & 82.9 & 29.2 \\
% \posl  & 90.7 & 50.4 & 87.4 & 31.0 \\ 
% \posh  & 91.3 & 56.4 & 91.7 & 34.7 \\ 
% \txtimg  & 92.0 & 81.3 & 93.0 & 41.7 \\ 
% \rowcolor{LightCyan}
% \genie{} (r=$0.5$) & 92.0 & 84.6 & 91.5 &  39.8\\
% \rowcolor{LightCyan}
% \genie{} (r=$0.6$) & 92.2 & 87.1 & 92.5 & 45.0 \\
% \rowcolor{LightCyan}
% \genie{} (r=$0.7$) & 92.5 & \textbf{87.9} & 92.9 & \textbf{47.0} \\
% \rowcolor{LightCyan}
% \genie{} (r=$0.8$) & 92.5 & 87.7 & \textbf{93.1} & 46.5 \\ 
% \rowcolor{LightCyan}
% \genie{} (r=$0.9$) & 92.4 & 87.1 & \textbf{93.1} &  45.7\\
% \rowcolor{LightCyan}
% \genienoad{} & \textbf{92.6} & \textbf{87.9} &  \textbf{93.1} & 46.9 \\
% \bottomrule
% \end{tabular}}

    \end{minipage}%
\vspace{-0.1in}
\end{table*}

\subsection{Fine-grained Few-shot Classification}
\label{sec:finegrained}

% chat GPT version 
To further investigate the impact of the proposed method, we compare \genie{} with other text-based data augmentation techniques across four distinct fine-grained datasets in a $20$-way, $1$-shot classification setting. We employ the pre-trained DINOV2 ViT-G \citep{oquab2023dinov2} backbone as a feature extractor to derive features from training images. Subsequently, an SVM classifier is trained on these features, and we report the Top-$1$ accuracy of the model on the test set.

\textbf{Datasets:}
We assess our method on several datasets: Food101 \citep{food101} with $101$ classes of foods, CUB200~\citep{Wah11thecaltech-ucsd} with $200$ bird species classes, Cars196 \citep{carsdataset} with $196$ car model classes, and FGVC-Aircraft \citep{aircraft} with $41$ aircraft manufacturer classes. We provide detailed information around fine-grained datasets in Table~\ref{tab:appendix_transfer_dset_details}. The reported metric is the average Top-$1$ accuracy over $100$ episodes. Each episode involves sampling $20$ classes and $1$-shot from the training set, with the final model evaluated on the respective test set.

\textbf{Implementation Details:}
We enhance the basic prompt by incorporating the superclass name for the fine-grained dataset: ``A photo of a $\small{<}\texttt{target class}\small{>}$, a type of $\small{<}\texttt{superclass}\small{>}$". For instance, in the \textit{food} dataset and the \textit{burger} class, our prompt reads: ``A photo of a \textit{burger}, a type of \textit{food}." No additional augmentation is used for generative methods in this context. We generate $19$ samples for both cases of our method and also the baseline with weak augmentation. 

\begin{wraptable}{R}{0.6\textwidth}
% \begin{table}[h!]
\vspace{-0.2in}
\caption{
\label{tab:finegrained} \textbf{Few-shot Learning on Fine-grained dataset:} We utilize an SVM classifier trained atop the DINOV2 ViT-G pretrained backbone, reporting Top-1 accuracy for the test set of each dataset. The baseline is an SVM trained on the same backbone using weak augmentation. Across all datasets, \genie{} surpasses this baseline.}

% \label{tab:finegrained} \textbf{Few-shot Learning on Finegrained dataset:} We train SVM classifier on top of DinoV2 ViT-G pretrained backbone and report Top-1 accuracy on the test set of each dataset. Our baseline here is the SVM trained with traditional augmentations. \genie{} outperforms Baseline across all dataset. Compared to data augmentation with \txtimg{} only, \genie{} improves accuracy by $6.6$ point and $5.2$ point in Cars and Aircraft dataset respectively. 

\centering
\scalebox{0.8}{
\begin{tabular}{lcccc}
\toprule 

\multicolumn{1}{l}{Method} & Birds  & Cars & Foods & Aircraft \\

\multicolumn{1}{l}{} &  \small{CUB200 \cite{Wah11thecaltech-ucsd}} & \small{Cars196 \cite{carsdataset}}  & \small{Food101 \cite{food101}} &  \small{Aircraft \cite{aircraft}} \\ \midrule
Baseline & 90.3 & 49.8 & 82.9 & 29.2 \\
\posl \cite{luzi2022boomerang}  & 90.7 & 50.4 & 87.4 & 31.0 \\ 
\posh \cite{luzi2022boomerang}  & 91.3 & 56.4 & 91.7 & 34.7 \\ 
\txtimg \cite{he2022synthetic}  & 92.0 & 81.3 & 93.0 & 41.7 \\ 
\rowcolor{LightCyan}
\genie{} (r=$0.5$) & 92.0 & 84.6 & 91.5 &  39.8\\
\rowcolor{LightCyan}
\genie{} (r=$0.6$) & 92.2 & 87.1 & 92.5 & 45.0 \\
\rowcolor{LightCyan}
\genie{} (r=$0.7$) & 92.5 & \textbf{87.9} & 92.9 & \textbf{47.0} \\
\rowcolor{LightCyan}
\genie{} (r=$0.8$) & 92.5 & 87.7 & \textbf{93.1} & 46.5 \\ 
\rowcolor{LightCyan}
\genie{} (r=$0.9$) & 92.4 & 87.1 & \textbf{93.1} &  45.7\\
\rowcolor{LightCyan}
\genienoad{} & \textbf{92.6} & \textbf{87.9} &  \textbf{93.1} & 46.9 \\
\bottomrule
\end{tabular}}

% \end{table}
\end{wraptable}

\textbf{Results:} Table~\ref{tab:finegrained} summarizes the results. \genie{} outperforms all other baselines, including \txtimg, by margins upto $0.5\%$ on CUB200 \citep{Wah11thecaltech-ucsd}, $6.6\%$ on Cars196 \citep{carsdataset}, $0.1\%$ on Food101 \citep{food101} and $5.3\%$ on FGVC-Aircraft \citep{aircraft}. Notably, \genie{} exhibits great effectiveness in more challenging datasets, outperforming the baseline with traditional augmentation by about $38\%$ for the Cars dataset and by roughly $17\%$ for the Aircraft dataset. It can be observed here that \genienoad{} performs on-par with \genie{} with a fixed noise level, eliminating the necessity for noise level search in \genie{}.
% \soroush{explain why \genie{} and \genienoad{} are on-par}
% \genienoad{} has on-par results with \genie{}. We hypothesis that is duo to the fact that semantic shift happens in ratio of $r \in (0.7,0.8)$, therefore 

\subsection{Ablation and Further Analysis}
\label{sec: abl_ana}

\begin{figure}[!t]
\centering

\includegraphics[width=1.0\linewidth]{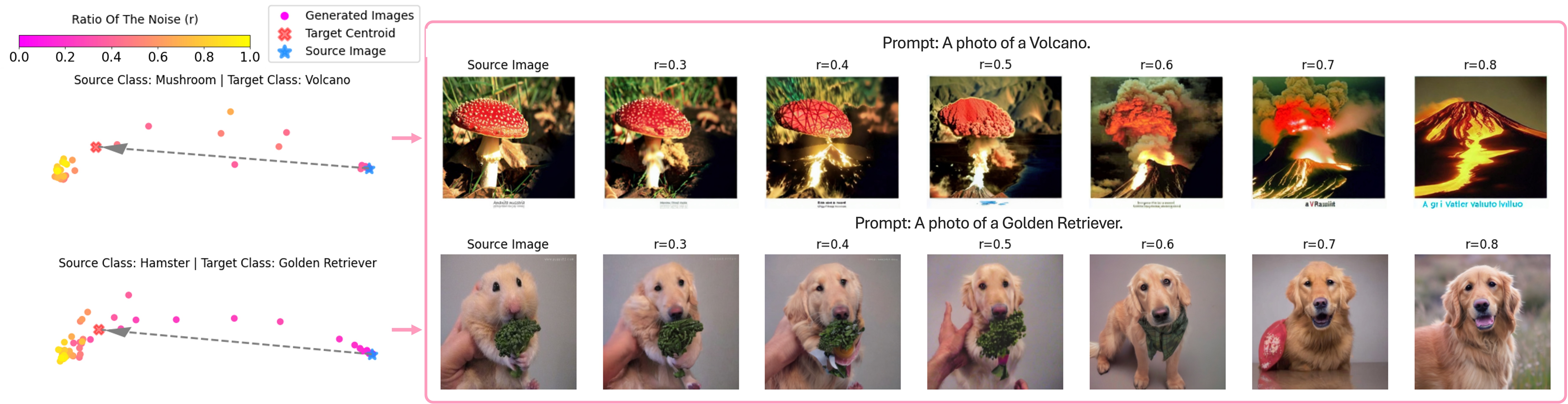}

\caption{\label{fig:traj} \textbf{Embedding visualizations of generative augmentations:} We pass all generative augmentations through DINOv2 ViT-G (serving as an oracle) to extract their corresponding embeddings and visualize them with PCA. As shown, the extent of semantic shifts varies based on both the source image and the target class. }

\end{figure}

\noindent \textbf{Semantic Shift from Source to Target Class.} The core motivation behind \genienoad{} is that by varying the noise ratio $r$ from $0$ to $1$, augmented sample $X_r$ will progressively shift its semantic category from source ($S$) in the beginning to target category ($T$) towards the end. However, somewhere between $0$ and $1$, $X_r$ will undergo a rapid transition from $S$ to $T$. To demonstrate this hypothesis empirically, in Figs.~\ref{fig:traj} and \ref{fig:traj_supp}, we visualize pairs of source images and target categories with their respective \genie{} generated augmentations for different noise ratios $r$, along with their corresponding PCA-projected embedding scatter plots (on the far left). We extract embeddings for all the images using a DINOv2 ViT-G pretrained backbone, which we assume as an oracle model in identifying the right category. We observe that as $r$ increases from $0.3$ to $0.8$, the images transition to embody more of the target category's semantics while preserving the contextual features of the source image. This transition of semantics can also be observed in the embedding plots (on the left) where they consistently shift from the proximity of the source image (blue star) to the target class's centroid (red cross) as the noise ratio $r$ increases. 
% Note that since embeddings for larger $r$ are closer to the target category, we call them \textit{hard negatives} for source category or \textit{hard positives} for target category. 
The sparse distribution of points within $r=[0.4,0.6]$ for the first image and $r=[0.2,0.4]$ for the second image aligns with our intuition of a rapid transition from category $S$ to $T$, thus empirically affirming our motivation behind \genienoad. 

\begin{figure}[!t]
\centering

\includegraphics[width=1.0\linewidth]{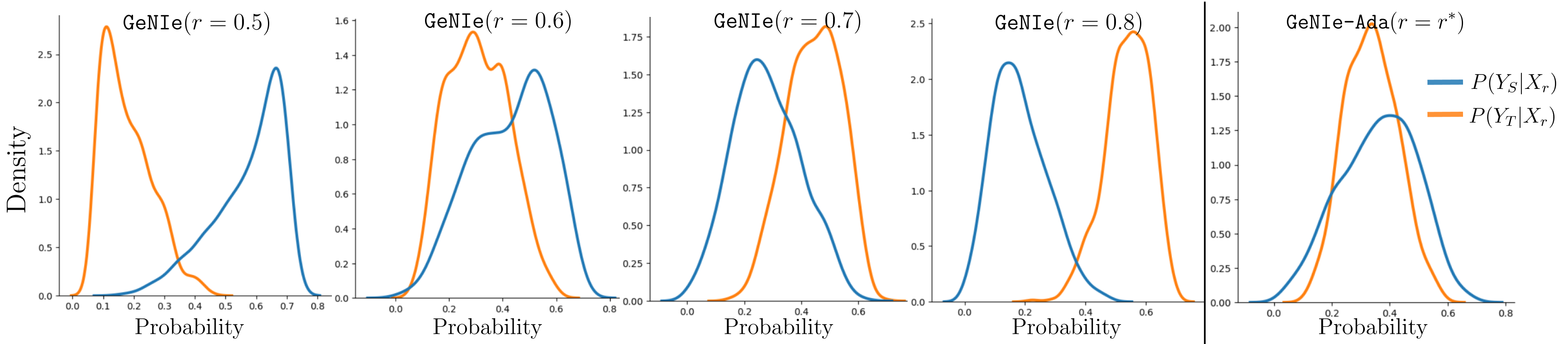}

\caption{\label{fig:prob} \textbf{Why \genie{} augmentations are challenging?} While deciding which class the generated augmentations ($X_r$) belong to is already difficult within $r=[0.6, 0.7]$ (due to high overlap between $P(Y_S|X_r)$ and $P(Y_T|X_r)$), \genienoad{} selects the best noise threshold ($r^*$) offering the hardest negative sample.}
%On average, the classifier confidently predicts the source class more than the target class for $X_r$ for $r=0.5$, and vice-versa for $r=0.8,0.9$. However, for $r=0.6,0.7$ and $r=r^*$ for \genienoad{}, the classifier struggles to classify $X_r$, indicating that the augmented samples are located closer to the decision boundary.}
\end{figure}

To further establish this, in Fig.~\ref{fig:prob}, we demonstrate the efficacy of \genie{} in generating hard negatives at the decision boundaries of an SVM classifier, which is trained on the labelled support set of the few-shot tasks of \textit{mini}-Imagenet, without any augmentations. We then plot  source and target class probabilities ($P(Y_S|X_r)$ and $P(Y_T|X_r)$, respectively) of the generated augmentation samples $X_r$. For both $r = 0.6$ and $0.7$, there is significant overlap between $P(Y_S|X_r)$ and $P(Y_T|X_r)$, making it difficult for the classifier to decide the correct class. On the right-hand-side, \genienoad{} automatically selects the best $r$ resulting in the most overlap between the two distributions, thus offering the hardest negative sample among the considered $r$ values (for more details see \ref{sec:classprob}). Note that a large overlap between distributions is not sufficient to call the generated samples hard negatives because they should also belong to the target category. This is, however, confirmed by the high Oracle accuracy in Table \ref{tab:noise_ablation} (elaborated in detail in the following paragraph) which verifies that majority of the generated augmentation samples do belong to the target category.

\noindent \textbf{Label consistency of the generated samples.} 
The choice of noise ratio $r$ is important in producing hard negative examples. In Table~\ref{tab:noise_ablation}, we present the accuracy of the \genie{} model across various noise ratios, alongside the oracle accuracy, which is an ImageNet pre-trained DeiT-Base \citep{touvron2021training} classifier. We observe a decline in the label consistency of generated data (quantified by the performance of the oracle model) when decreasing the noise level. Reducing $r$ also results in a degradation in the performance of the final few-shot model ($87.2\% \rightarrow 77.6\%$) corroborating that an appropriate choice of $r$ plays a crucial role in our design strategy. We investigate this further in the following paragraph.

\noindent \textbf{Effect of Noise in \genie.} 
 We examine the impact of noise on the performance of the few-shot model in Table~\ref{tab:noise_ablation}. Noise levels $r \in [0.7, 0.8]$ yield the best performance. Conversely, utilizing noise levels below $0.7$ diminishes performance due to label inconsistency, as is demonstrated in Table~\ref{tab:noise_ablation} and Fig~\ref{fig:traj}. As such, determining the appropriate noise level is pivotal for the performance of \genie{} to be able to generate challenging hard negatives while maintaining label consistency. An alternative approach to finding the optimal noise level involves using \genienoad{} to adaptively select the noise level for each source image and target class. As demonstrated in Tables~\ref{tab:noise_ablation} and \ref{tab:finegrained}, \genienoad{} achieves performance that is comparable to or surpasses that of \genie{} with fixed noise levels.

\noindent \textbf{Effect of Diffusion Models in \genie.} 
We have tried experimenting with both smaller as well as more recent diffusion models. More specifically, we have used Stable Diffusion XL-Turbo to generate hard-negatives through $\texttt{GeNIe}$ and $\texttt{GeNIe-Ada}$. Few-shot classification results on miniImagenet with these augmentations are shown in Table~\ref{tab:noise_ablation}. The accuracies follow a similar trend to that of Table~\ref{tab:mini_imagenet}, where Stable Diffusion 1.5 was used to generate augmentations. $\texttt{GeNIe-Ada}$ improves UniSiam's few-shot performance the most as compared to $\texttt{GeNIe}$ with different noise ratios $r$, and even when compared to $\texttt{Txt2Img}$. This empirically indicates the robustness of $\texttt{GeNIe}$ and $\texttt{GeNIe-Ada}$ to different diffusion engines. Note that, Stable Diffusion XL-Turbo by default uses $4$ steps for the sake of optimization, and to ensure we can have the right granularity for the choice of $r$ we have set the number of steps to $10$. That is already 5 times faster than the standard Stable Diffusion v1.5 with $50$ steps. Our experiments with Stable Diffusion v3 (which is a totally different model with a Transformers backbone) also in Table~\ref{tab:noise_ablation} also convey the same message. As such, we believe our approach is generalizable across different diffusion models.  

% To address this, one possible approach could be to leverage additional information from generated images during training as soft labels (e.g., soft cross-entropy loss or ranking loss). We leave the exploration of this avenue to future research.\\

% \noindent \textbf{Why are the generated samples Hard Negatives? } 
% Given that the generated images share visual features with the source image, we posit that they should be situated in close proximity to the source image in the embedding space while embodying the semantics of the target class. To provide further insight into our hypothesis, we generate UMAP \citep{mcinnes2018umap} plots for augmentation samples from various generative methods, as illustrated in Fig~\ref{fig:diff_augs_comp}. Initially, we note that samples generated by \genie{} exhibit a more dispersed distribution, spanning the boundaries of classes compared to other augmentation methods. We characterize these samples as hard negatives. Additionally, a reduction in the noise level within \genie{} results in more diverse samples, even though this diversity comes with a trade-off in label consistency, as discussed earlier.\\

\begin{table}[!t]
 \centering
\caption{\label{tab:noise_ablation} \textbf{Effect of Noise and Diffusion Models in \genie:} We use the same setting as in Table \ref{tab:mini_imagenet} to study the effect of the amount of noise. As expected (also shown in Fig~\ref{fig:traj}), small noise results in worse accuracy since some generated images may be from the source category rather than the target one. For $r=0.5$ only $73\%$ of the generated data is from the target category. This behaviour is also shown in Fig.~\ref{fig:noise_ab}. Notably, reducing the noise level below $0.7$ is associated with a decline in oracle accuracy and subsequent degradation in the performance of the final few-shot model. Note that the high oracle accuracy of \genienoad{} demonstrates its capability to adaptively select the noise level per source and target, ensuring semantic consistency with the intended target. 
To further demonstrate \genie{}'s ability to generalize across different diffusion models, we replace the diffusion model with SD3 and SDXL-Turbo. The resulting accuracies follow a similar trend to those in Table~\ref{tab:mini_imagenet}, confirming \genie{}'s advantage over \txtimg{} across various diffusion models.}
\vspace{0.05in}
 \setlength{\tabcolsep}{3.0mm}{
{
 \scalebox{0.8}{
  \begin{tabular}{lcc|cc|cc|cc|c}
    \toprule
    
    \textbf{Method} & \textbf{Generative} & \textbf{Noise}  & \multicolumn{2}{c|}{\textbf{ResNet-18}} &  \multicolumn{2}{c|}{\textbf{ResNet-34}}& \multicolumn{2}{c|}{\textbf{ResNet-50}}  & \textbf{Oracle}\\
    & \textbf{Model} & \textbf{r=} & \textbf{1-shot} & \textbf{5-shot} & \textbf{1-shot} & \textbf{5-shot} & \textbf{1-shot} & \textbf{5-shot}& \textbf{Acc}\\
    \midrule
    \txtimg{} & SD 1.5 & - & 74.1$\pm$0.6 & 84.6$\pm$0.5 & 75.4$\pm$0.6 & 85.5$\pm$0.5 & 76.4$\pm$0.6 & 86.5$\pm$0.4 & -\\
     \genie{} & SD 1.5 & 0.5 & 60.4$\pm$0.8 & 74.1$\pm$0.6 & 62.0$\pm$0.8 & 75.8$\pm$0.6 & 63.7$\pm$0.9 & 77.6$\pm$0.6 & 73.4$\pm$0.5\\
     \genie{} & SD 1.5 & 0.6 & 69.7$\pm$0.7 & 80.7$\pm$0.5 & 71.1$\pm$0.7 & 82.2$\pm$0.5 & 72.1$\pm$0.7 & 82.8$\pm$0.5 & 85.8$\pm$0.4\\
     \genie{} & SD 1.5 & 0.7 & 74.5$\pm$0.6 & 83.3$\pm$0.5 & 76.4$\pm$0.6 & 84.4$\pm$0.5 & 77.1$\pm$0.6 & 85.0$\pm$0.4 & 94.5$\pm$0.2\\
     \genie{} & SD 1.5 & 0.8 & 75.5$\pm$0.6 & 85.4$\pm$0.4 & 77.1$\pm$0.6 & 86.3$\pm$0.4 & 77.3$\pm$0.6 & 87.2$\pm$0.4 & 98.2$\pm$0.1\\
     \genie{} & SD 1.5 & 0.9 & 75.0$\pm$0.6 & 85.3$\pm$0.4 & 77.6$\pm$0.6 & 86.2$\pm$0.4 & 77.7$\pm$0.6 & 87.0$\pm$0.4 & 99.3$\pm$0.1\\
     \rowcolor{LightCyan}
     \genienoad{} & SD 1.5 & Adaptive & 76.8$\pm$0.6 & 85.9$\pm$0.4 & 78.5$\pm$0.6 & 86.6$\pm$0.4 & 78.6$\pm$0.6 & 87.9$\pm$0.4 & 98.9$\pm$0.2\\
\midrule
     \txtimg{} & SDXL-Turbo & - & 72.5$\pm$0.3 & 82.1$\pm$0.6 & 76.2$\pm$0.2 & 84.4$\pm$0.3 & 76.7$\pm$0.6 & 85.9$\pm$0.5 & -\\
     \genie{} & SDXL-Turbo & 0.5 & 61.2$\pm$0.5 & 73.5$\pm$0.2 & 61.5$\pm$0.2 & 74.9$\pm$0.3 & 63.1$\pm$0.2 & 76.5$\pm$0.6 & - \\
     \genie{} & SDXL-Turbo & 0.6 & 70.2$\pm$0.2 & 79.3$\pm$0.4 & 71.2$\pm$0.7 & 81.4$\pm$0.6 & 73.2$\pm$0.2 & 82.4$\pm$0.5 & - \\
     \genie{} & SDXL-Turbo & 0.7 & 73.1$\pm$0.3 & 83.5$\pm$0.5 & 76.1$\pm$0.6 & 85.3$\pm$0.4 & 77.2$\pm$0.6 & 84.2$\pm$0.4 & - \\
     \genie{} & SDXL-Turbo & 0.8 & 74.2$\pm$0.3 & 85.1$\pm$0.3 & 76.9$\pm$0.4 & 85.5$\pm$0.5 & 78.7$\pm$0.6 & 87.7$\pm$0.4 & - \\
     \genie{} & SDXL-Turbo & 0.9 & 73.9$\pm$0.4 & 84.9$\pm$0.7 & 76.6$\pm$0.7 & 84.2$\pm$0.6 & 78.1$\pm$0.5 & 87.0$\pm$0.4 & - \\
     \rowcolor{LightCyan}
     \genienoad{} & SDXL-Turbo & Adaptive & 75.1$\pm$0.3 & 87.1$\pm$0.8 & 78.9$\pm$0.5 & 85.2$\pm$0.5 & 79.0$\pm$0.6 & 88.6$\pm$0.2 & - \\
     \midrule
     \txtimg{} & SD 3 & - & 73.6$\pm$1.7 & 82.9$\pm$1.2 & 76.7$\pm$1.5 & 85.5$\pm$1.3 & 77.2$\pm$1.9 & 85.0$\pm$1.2 & - \\
     \genie{} & SD 3 & 0.5 & 62.0$\pm$1.2 & 72.9$\pm$1.1 & 62.5$\pm$0.9 & 73.9$\pm$1.0 & 64.1$\pm$0.5 & 76.1$\pm$1.9 & - \\
     \genie{} & SD 3 & 0.6 & 70.8$\pm$1.5 & 79.1$\pm$1.9 & 71.8$\pm$1.2 & 82.1$\pm$1.3 & 74.1$\pm$1.5 & 83.4$\pm$1.8 & - \\
     \genie{} & SD 3 & 0.7 & 74.6$\pm$0.8 & 84.5$\pm$1.2 & 76.5$\pm$1.9 & 86.2$\pm$1.6 & 78.5$\pm$1.9 & 84.0$\pm$1.1 & - \\
     \genie{} & SD 3 & 0.8 & 75.9$\pm$1.2 & 86.3$\pm$1.7 & 77.8$\pm$1.9 & 85.5$\pm$1.9 & 79.2$\pm$1.7 & 88.3$\pm$1.9 & - \\
     \genie{} & SD 3 & 0.9 & 75.1$\pm$0.5 & 85.2$\pm$1.2 & 78.1$\pm$1.3 & 86.2$\pm$1.2 & 77.1$\pm$1.9 & 88.9$\pm$0.8 & - \\
     \rowcolor{LightCyan}
     \genienoad{} & SD 3 & Adaptive & 76.8$\pm$1.3 & 87.5$\pm$1.5 & 78.9$\pm$1.3 & 87.7$\pm$1.5 & 79.1$\pm$1.4 & 89.5$\pm$1.0 & - \\
    \bottomrule
    
  \end{tabular}}
}
    }
    \vspace{-0.2in}
\end{table}

\vspace{-0.1in}
\section{Concluding Remarks}
\vspace{-0.1in}
\genie{}, for the first time to our knowledge, combines contradictory sources of information (a source image, and a different target category prompt) through a noise adjustment strategy into a conditional latent diffusion model to generate challenging augmentations, which can serve as hard negatives. 

\noindent \textbf{Limitation.}
\label{sec:limitation}
The required time to create augmentations through \genie{} is on par with any typical diffusion-based competitors \citep{azizi2023synthetic,he2022synthetic}; however, this is naturally slower than traditional augmentation techniques \citep{Cutmix,mixup}. This is not a bottleneck in offline augmentation strategies, but can be considered a limiting factor in real-time scenarios. Recent studies are already mitigating this through advancements in diffusion model efficiency  \citep{sauer2023adversarial,meng2023distillation,liu2023instaflow}. Another challenge present in any generative AI-based augmentation technique is the domain shift between the distribution of training data and the downstream context they might be used for augmentation. A possible remedy is to fine-tune the diffusion backbone on a rather small dataset from the downstream task.

%Furthermore, we anticipate that \genie{} may encounter challenges when applied to datasets where the images deviate from the distribution expected by the generative backbone or when category names are unfamiliar to the text encoder within the generative method. It's important to note that this limitation extends to any augmentation methods rooted in text-to-image generative approaches. A potential resolution lies in fine-tuning the text-to-image generative model on a small set of image-text pairs relevant to the downstream task. By doing so, the generative model becomes more attuned to the unique characteristics of the task, enhancing its effectiveness when employed as a data augmentation technique.% in the subsequent task.

% We measure the time for augmentation of each images. Since augmenting with different noise level requires different number of reverse diffusion iterations, timing varies depend on the noise level. For images with 512 by 512 resolution, augmenting each image takes $4.4$ second for noise level of $r=0.5$ and $6.7$ second for noise level of $r=0.8$. Note that this is much higher compare to traditional augmentation methods (e.g., cropping and resizing) and limit the application of \genie{} to online augmentation of images. However, this limitation can be mitigate with further advancement in the efficiency of diffusion models. For example, recent diffusion models \citep{} significantly improve the runtime of diffusion models compared to Stable Diffusion 1.5. Note that we did not evaluate these models for \genie{} as it is out of scope of our paper. 

\noindent \textbf{Broader Impact.} We believe ideas from \genie{} can have a significant impact when it comes to generating hard augmentations challenging and thus enhancing downstream tasks beyond classification. At the same time, just like any other generative model, \genie{} can also introduce inherent biases stemming from the training data used to build its diffusion backbone, which can reflect and amplify societal prejudices or inaccuracies. Therefore, it is crucial to carefully mitigate potential biases in generative models such as \genie{} to ensure a fair and ethical deployment of deep learning systems.

%\genie{} leverages diffusion models to generate challenging samples for the target category while retaining some low-level and contextual features from the source image. Our experiments, spanning few-shot and long-tail distribution settings, showcase \genie{}'s effectiveness, especially in categories with limited examples. 
% We discuss limitation of our method in Sec.~\ref{sec:limitation}. 

% \noindent\textbf{Acknowledgments.}
% This work is partially funded by NSF grant number 1845216, DARPA Contract No. HR00112290115, and 
% Shell.ai Innovation Program at Shell Global Solutions International B.V.

\bibliography{tmlr}
\bibliographystyle{tmlr}

\setcounter{figure}{0}
\renewcommand\thefigure{A\arabic{figure}}
\setcounter{table}{0}
\renewcommand\thetable{A\arabic{table}}
\appendix
\onecolumn
\vspace{-2.0in}
\section{Appendix}

\subsection{Analyzing \genie{}, \genienoad{}'s Class-Probabilities}
\label{sec:classprob}
The core aim of \genie{} and \genienoad{} is to address the failure modes of a classifier by generating \emph{challenging} samples located near the decision boundary of each class pair, which facilitates the learning process in effectively enhancing the decision boundary between classes. As summarized in Table~\ref{tab:noise_ablation} and illustrated in Fig.~\ref{fig:traj}, we have empirically corroborated that \genie{} and \genienoad{} can respectively produce samples $X_r, X_{r^{*}}$ that are negative with respect to the source image $X_S$, while semantically belonging to the class $T$.
% \begin{figure}[h!]
% \centering
% \includegraphics[width=1.0\linewidth]{ICLR 2025 Template/figures/genie-p.png}
% %\vspace{-.05in}
% \caption{$P(Y_S|X_r)$ and $P(Y_T|X_r)$ for $r \in \{0.5,0.6,0.7,0.8,0.9\}$. On average, the classifier confidently predicts the source class more than the target class for $X_r$ for $r=0.5$, and vice-versa for $r=0.8,0.9$. However, for $r=0.6,0.7$, the classifier struggles to classify $X_r$, indicating that the augmented samples are located closer to the decision boundary.}
% \label{fig:genie-p}
% %\vspace{-0.2in}
% \end{figure}
To further analyze the effectiveness of \genie{} and \genienoad{}, we compare the source class-probabilities $P(Y_S|X_r)$ and target-class probabilities $P(Y_S|X_r)$ of  augmented samples $X_r$. 
\begin{wrapfigure}{r}{0.3\textwidth}
  \begin{center}
    \includegraphics[width=0.4\textwidth]{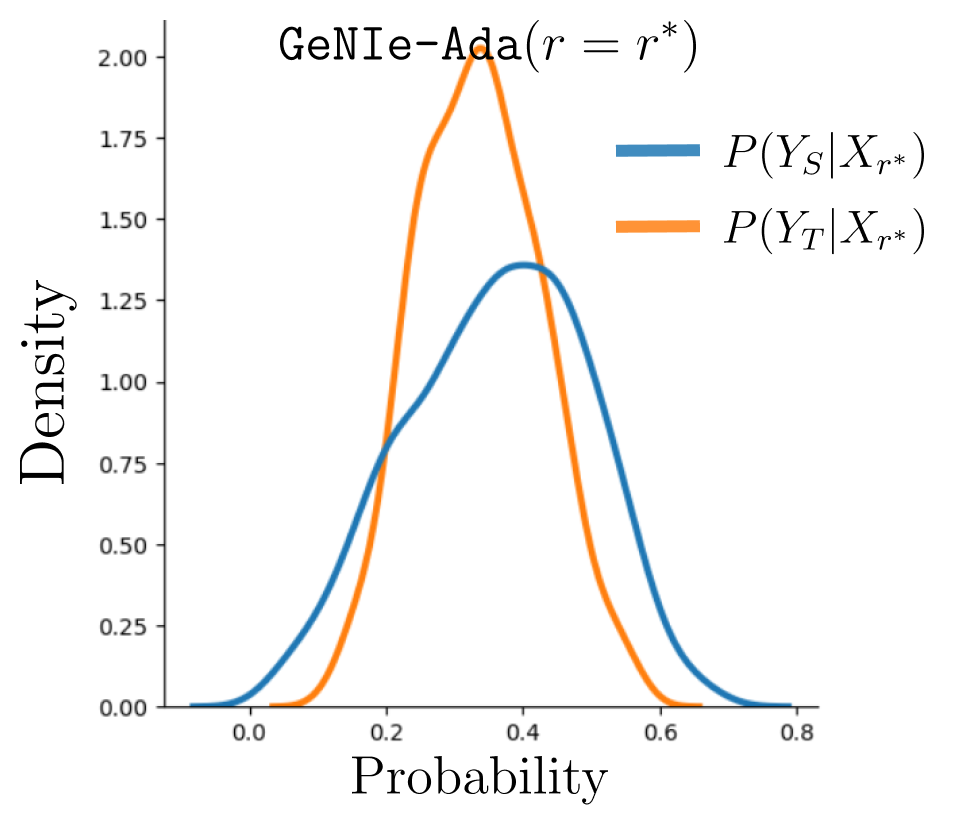}
  \end{center}
  \caption{Significant overlap between $P(Y_S|X_{r^{*}})$ and $P(Y_T|X_{r^{*}})$ indicates high class-confusion for augmented samples generated by \genienoad{}.}
  \label{fig:genie-ada-p}
\end{wrapfigure}
To compute these class probabilities, we first fit an SVM classifier (as followed in UniSiam \citep{unisiam}) only on the labelled support set embeddings of each episode in the \textit{mini}Imagenet test dataset. Then, we perform inference using each episode's SVM classifier on its respective $X_r$'s and extract its class probabilities of belonging to its source class $S$ and target class $T$. These per augmentation-sample source and target class probabilities are then averaged for each episode for each $r \in \{0.5,0.6,0.7,0.8,0.9\}$ in the case of \genie{} and for the optimal $r = r^{*}$ per sample in the case of \genienoad{}, plotted as density plots in Fig.~\ref{fig:prob}, Fig.~\ref{fig:genie-ada-p}, respectively. Fig.~\ref{fig:prob} illustrates that $P(Y_S|X_r)$ and $P(Y_T|X_r)$ have significant overlap in the case of $r \in \{0.6,0.7\}$ indicating class-confusion for $X_r$.
% whereas $r \in \{0.5,0.8,0.9\}$ show significant separation between $P(Y_S|X_r)$ and $P(Y_T|X_r)$. 

% \begin{figure}[h!]
% \centering
% \includegraphics[width=0.28\linewidth]{figures/genie-ada-p.png}
% %\vspace{-.05in}
% \caption{}
% \label{fig:genie-ada-p}
% %\vspace{-0.2in}
% \end{figure}

Furthermore, Fig.~\ref{fig:genie-ada-p} illustrates that when using the optimal $r = r^{*}$ found by \genienoad{} per sample, $P(Y_S|X_r)$ and $P(Y_T|X_r)$ significantly overlap around probability scores of $0.2-0.45$, indicating class confusion for \genienoad{} augmentations. This corroborates with our analysis in Section \ref{sec: abl_ana}, Table~\ref{tab:noise_ablation} and additionally empirically proves that the augmented samples generated by \genie{} for $r \in \{0.6,0.7\}$ and \genienoad{} for $r = r^{*}$ are actually located near the decision boundary of each class pair.

\begin{table} [h!]
    \caption{Train and test split details of the fine-grained datasets. We use the provided train set for few-shot task generation, and the provided test sets for our evaluation. Aircraft dataset uses the manufacturer hierarchy.}
    \vspace{-0.1in}
    \label{tab:appendix_transfer_dset_details}
    \begin{center}
    \scalebox{0.70}{
    \begin{tabular}{lcccccc}
        \toprule
        Dataset & Classes & Train  & Test  \\
         &  &  samples &  samples \\
        \midrule
        CUB200 \citep{Wah11thecaltech-ucsd} & 200 & 5994  & 5794 \\
        Food101 \citep{food101} & 101 & 75750 & 25250 \\
        Cars \citep{carsdataset} & 196 & 8144 & 8041 \\
        Aircraft \citep{aircraft} & 41 & 6,667  & 3333\\

        \bottomrule
    \end{tabular}
    }

    \end{center}
    \vspace{-0.3in}
\end{table}

% \vspace{0.05cm}
%         \caption{\label{tab:finegrained} \textbf{Few-shot Learning on Fine-grained dataset:} We utilize a SVM classifier trained atop the DinoV2 ViT-G pretrained backbone, reporting Top-1 accuracy for the test set of each dataset. The baseline is an SVM trained on the same backbone using weak augmentation. Across all datasets, \genie{} surpasses this baseline.}
%         % \vspace{-0.25cm}
%         % Compared to data augmentation with \txtimg{} only, \genie{} improves accuracy by $6.6$ point and $5.2$ point in Cars and Aircraft dataset respectively. }
% \scalebox{0.63}{
% \begin{tabular}{lcccc}
% \toprule 

% % \multicolumn{1}{l}{Method} & Birds   & Cars & Foods & Aircraft \\

% \multicolumn{1}{l}{Method} &  \small{CUB200 \citep{Wah11thecaltech-ucsd}} & \small{Cars196 \citep{carsdataset}}  & \small{Food101 \citep{food101}} &  \small{Aircraft \citep{aircraft}} \\ \midrule
% Baseline & 90.3 & 49.8 & 82.9 & 29.2 \\
% \posl  & 90.7 & 50.4 & 87.4 & 31.0 \\ 
% \posh  & 91.3 & 56.4 & 91.7 & 34.7 \\ 
% \txtimg  & 92.0 & 81.3 & 93.0 & 41.7 \\ 
% \rowcolor{LightCyan}
% \genie{} (r=$0.5$) & 92.0 & 84.6 & 91.5 &  39.8\\
% \rowcolor{LightCyan}
% \genie{} (r=$0.6$) & 92.2 & 87.1 & 92.5 & 45.0 \\
% \rowcolor{LightCyan}
% \genie{} (r=$0.7$) & 92.5 & \textbf{87.9} & 92.9 & \textbf{47.0} \\
% \rowcolor{LightCyan}
% \genie{} (r=$0.8$) & 92.5 & 87.7 & \textbf{93.1} & 46.5 \\ 
% \rowcolor{LightCyan}
% \genie{} (r=$0.9$) & 92.4 & 87.1 & \textbf{93.1} &  45.7\\
% \rowcolor{LightCyan}
% \genienoad{} & \textbf{92.6} & \textbf{87.9} &  \textbf{93.1} & 46.9 \\
% \bottomrule
% \end{tabular}}

\subsection{Few-shot Classification with ResNet-34 on \textit{tiered}Imagenet}
\label{sec:r34tiered}
\begin{table}[h!]
    % \caption{Global caption}
        \caption{\label{tab:r34t} \textbf{\textit{tiered-}ImageNet:} Accuracies (\% $\pm$ std) for $5$-way, $1$-shot and $5$-way, $5$-shot classification settings on the test-set. We compare against various SOTA supervised and unsupervised few-shot classification baselines as well as other augmentation methods, with UniSiam \citep{unisiam} pre-trained ResNet-34 backbone.}
        
      \centering
        \scalebox{0.8}{
        \begin{tabular}{llccc}
    \toprule
    \multicolumn{5}{c}{\textbf{ResNet-34}} \\ \midrule
    \textbf{Augmentation} & \textbf{Method}  & \textbf{Pre-training} & \textbf{1-shot} & \textbf{5-shot}  \\
    \midrule
    Weak & MAML + dist \citep{MAML}   & sup. & 51.7$\pm$1.8 & 70.3$\pm$1.7 \\
    Weak & ProtoNet \citep{ProtoNet}   & sup. & 52.0$\pm$1.2 & 72.1$\pm$1.5 \\
    \midrule
    Weak & UniSiam + dist \citep{unisiam}   & unsup. & 68.7$\pm$0.4 & 85.7$\pm$0.3 \\
    Weak & UniSiam \citep{unisiam}   & unsup. & 65.0$\pm$0.7 & 82.5$\pm$0.5 \\
    Strong & UniSiam \citep{unisiam}  & unsup.  & 64.8$\pm$0.7 & 82.4$\pm$0.5 \\
    CutMix \citep{Cutmix} & UniSiam \citep{unisiam}  & unsup.  & 63.8$\pm$0.7 & 80.3$\pm$0.6 \\
    MixUp \citep{mixup} & UniSiam \citep{unisiam} & unsup.  & 64.1$\pm$0.7 & 80.0$\pm$0.6 \\
    \posl \citep{luzi2022boomerang} & UniSiam \citep{unisiam}  & unsup.  & 66.1$\pm$0.7 & 83.1$\pm$0.5 \\
    \posh \citep{luzi2022boomerang} & UniSiam \citep{unisiam} & unsup.  & 70.4$\pm$0.7 & 84.7$\pm$0.5 \\
    \txtimg \citep{he2022synthetic} & UniSiam \citep{unisiam} & unsup.  & 75.0$\pm$0.6 & 85.4$\pm$0.4 \\
    DAFusion \citep{trabucco2024effective} & UniSiam \citep{unisiam} & unsup.  & 64.1$\pm$2.1 & 82.8$\pm$1.4  \\
    \rowcolor{LightCyan}
    \genie{} (Ours) & UniSiam \citep{unisiam}  & unsup.  & \textbf{75.7$\pm$0.6} & \textbf{86.0$\pm$0.4} \\
     \rowcolor{LightCyan}
    \genienoad{} (Ours) & UniSiam \citep{unisiam}  & unsup.  & \textbf{76.9$\pm$0.6} & \textbf{86.3$\pm$0.2} \\
    
    \bottomrule
  \end{tabular}
        
        }
\end{table}
We follow the same evaluation protocol here as mentioned in section \ref{sec:few_shot}. As summarized in Table~\ref{tab:r34t}, \genie{} and \genienoad{} outperform all other data augmentation techniques. 

\subsection{Additional details of Long-Tail experiments}
\label{sec:appendix_longtail}
We present a comprehensive version of Table \ref{tab:imagenet_lt} to benchmark the performance with different backbone architectures (e.g., ResNet50) and to compare against previous long-tail baselines; this is detailed in Table \ref{tab:imagenet_lt_apendix}.

\textbf{Implementation Details of LViT:} We download the pre-trained ViT-B of LViT \citep{xu2023learning} and finetune it with Bal-BCE loss proposed therein on the augmented dataset. Training takes 2 hours on four NVIDIA RTX 3090 GPUs. We use the same hyperparameters as in \citep{xu2023learning} for finetuning: $100$ epochs, $lr=0.008$, batch size of $1024$, CutMix and MixUp for the data augmentation.   

\textbf{Implementation Details of VL-LTR:} We use the official code of VL-LTR \citep{tian2022vl} for our experiments. We use a pre-trained CLIP ResNet-50 backbone. We followed the hyperparameters reported in VL-LTR \citep{tian2022vl}. We augment only ``Few'' category and train the backbone with the VL-LTR \citep{tian2022vl} method. Training takes 4 hours on $8$ NVIDIA RTX 3090 GPUs.

\begin{table}[!t]
 \centering
\caption{\label{tab:ex_comput} Few-shot classification comparison of GeNIe-Ada with \txtimg{} on miniImagenet.}
% , evaluated on 200 episodes. We generate 4 augmentations per class for 1-shot and 20 per class for 5-shot settings using GeNIe-Ada, and 8 augmentations per class for 1-shot and 40 per class for 5-shot settings using \txtimg{} to keep a constant compute budget.}
 \setlength{\tabcolsep}{4.0mm}{
{
 \scalebox{0.8}{
 \small
  \begin{tabular}{l|cc|cc|cc}
    \toprule
    
    \textbf{Method}  & \multicolumn{2}{c|}{\textbf{ResNet-18}} &  \multicolumn{2}{c|}{\textbf{ResNet-34}}& \multicolumn{2}{c}{\textbf{ResNet-50}} \\
    & \textbf{1-shot} & \textbf{5-shot} & \textbf{1-shot} & \textbf{5-shot} & \textbf{1-shot} & \textbf{5-shot}\\
    \midrule
    \txtimg{} & 76.9$\pm$1.0 & 86.5$\pm$0.9 & 77.1$\pm$0.8 & 86.7$\pm$1.0 & 77.2$\pm$1.3 & 86.8$\pm$0.9\\
     \rowcolor{LightCyan}
     \genienoad{} & 77.7$\pm$0.8 & 87.4$\pm$1.0 & 78.3$\pm$0.9 & 87.8$\pm$0.9 & 79.1$\pm$1.1 & 88.4$\pm$1.2\\
    \bottomrule
    
  \end{tabular}}
}
    }
\end{table}
\vspace{-0.2cm}

\begin{table}[!b]
\caption{
\label{tab:imagenet_lt_apendix}
 \textbf{Long-Tailed ImageNet-LT:}  
 We compare different augmentation methods on ImageNet-LT and report Top-1 accuracy for ``Few'', ``Medium'', and ``Many'' sets. $\dagger$ indicates results with ResNeXt50. $*$: indicates training with 384 resolution so is not directly comparable with other methods with 224 resolution. On the ``Few'' set and LiVT method, our augmentations improve the accuracy by $11.7$ points compared to LiVT original augmentation and $4.4$ points compared to \txtimg{}.}
 
\centering
\scalebox{0.823}{
\begin{tabular}{lccc|c}

\toprule
\multicolumn{5}{c}{\cellcolor{pink}\textbf{ResNet-50}} \\\midrule
\multicolumn{1}{l}{Method}  & Many & Med.        & Few           & Overall Acc  \\ \midrule
CE \citep{CB} & 64.0 & 33.8   & 5.8  & 41.6 \\
LDAM \citep{LDAM} & 60.4 & 46.9   & 30.7 & 49.8 \\
c-RT \citep{NCM}  & 61.8 & 46.2   & 27.3 & 49.6 \\
$\tau$-Norm \citep{NCM}  & 59.1 & 46.9   & 30.7 & 49.4 \\
Causal \citep{CausalNorm}  & 62.7 & 48.8   & 31.6 & 51.8 \\
Logit Adj. \citep{LA} & 61.1 & 47.5   & 27.6 & 50.1 \\
RIDE(4E)$\dagger$ \citep{RIDE}  & 68.3 & 53.5   & 35.9 & 56.8 \\
MiSLAS \citep{MiSLAS}  & 62.9	& 50.7	 & 34.3	& 52.7 \\
DisAlign \citep{DisAlign}  & 61.3 & 52.2   & 31.4 & 52.9 \\
ACE$\dagger$ \citep{ACE} & 71.7 & 54.6   & 23.5 & 56.6 \\
PaCo$\dagger$ \citep{PaCo}  & 68.0 & 56.4   & 37.2 & 58.2 \\
TADE$\dagger$ \citep{TADE}  & 66.5 & \textbf{57.0}   & 43.5 & 58.8 \\
TSC \citep{TSC}  & 63.5 & 49.7   & 30.4 & 52.4 \\
GCL \citep{GCL}  & 63.0 & 52.7   & 37.1 & 54.5 \\
TLC \citep{TLC}  & 68.9 & 55.7   & 40.8 & 55.1 \\
BCL$\dagger$ \citep{BCL}  & 67.6 & 54.6   & 36.6 & 57.2 \\
NCL \citep{NCL}  & 67.3 & 55.4   & 39.0 & 57.7 \\
SAFA \citep{SAFA}  & 63.8 & 49.9   & 33.4 & 53.1 \\
DOC \citep{DOC} & 65.1 & 52.8   & 34.2 & 55.0 \\
DLSA \citep{DLSA}  & 67.8 & 54.5   & 38.8 & 57.5 \\ 
{ResLT~\citep{cui2022reslt}} & {63.3} & {53.3} & {40.3} & {55.1}\\
  {PaCo~\citep{cui2021parametric}}  & {68.2} & {58.7} & {41.0} & {60.0}\\
  {LWS~\citep{kang2019decoupling}}  & {62.2} & {48.6} & {31.8} & {51.5}\\
  {Zero-shot CLIP~\citep{radford2021learning}}  & {60.8} & {59.3} & {58.6} & {59.8}\\ 
 {DRO-LT~\citep{samuel2021distributional} } & {64.0} & {49.8} & {33.1} & {53.5}\\
 {VL-LTR~\citep{tian2022vl}}  & {77.8} & {67.0} & {50.8} & {70.1}\\
 Cap2Aug \citep{roy2023cap2aug}  & {78.5} & {\textbf{67.7}} & {51.9} & {70.9}\\  	
 \rowcolor{LightCyan}
 \genienoad{}  & {\textbf{79.2}} & {64.6} & {\textbf{59.5}} & {\textbf{71.5}}\\
 \toprule
\multicolumn{5}{c}{\cellcolor{pink}\textbf{ViT-B}} \\
\midrule
LiVT* \citep{xu2023learning} & 76.4 & 59.7  & 42.7 & 63.8 \\
\midrule
ViT \citep{ViT} &  50.5 & 23.5   & 6.9 & 31.6 \\
MAE \citep{MAE}  &  74.7 & 48.2   & 19.4 & 54.5 \\
DeiT \citep{deit3} &  70.4     & 40.9     & 12.8     & 48.4 \\
LiVT \citep{xu2023learning} &  73.6 & 56.4   & 41.0 & 60.9 \\
LiVT + \posl  & 74.3 & 56.4 & 34.3 & 60.5\\
LiVT + \posh  &  73.8 & 56.4 & 45.3 & 61.6\\
LiVT + \txtimg  &  \textbf{74.9} & 55.6 & 48.3 & 62.2\\
\rowcolor{LightCyan}
LiVT + \genie{} (r=$0.8$) &  74.5 & 56.7 & 50.9 & 62.8\\
\rowcolor{LightCyan}
LiVT + \genienoad{} &  74.0 & \textbf{56.9} & \textbf{52.7} & \textbf{63.1}\\
\bottomrule
\end{tabular}}

\end{table}

\begin{figure}[!t]
\centering
\includegraphics[width=0.9\linewidth]{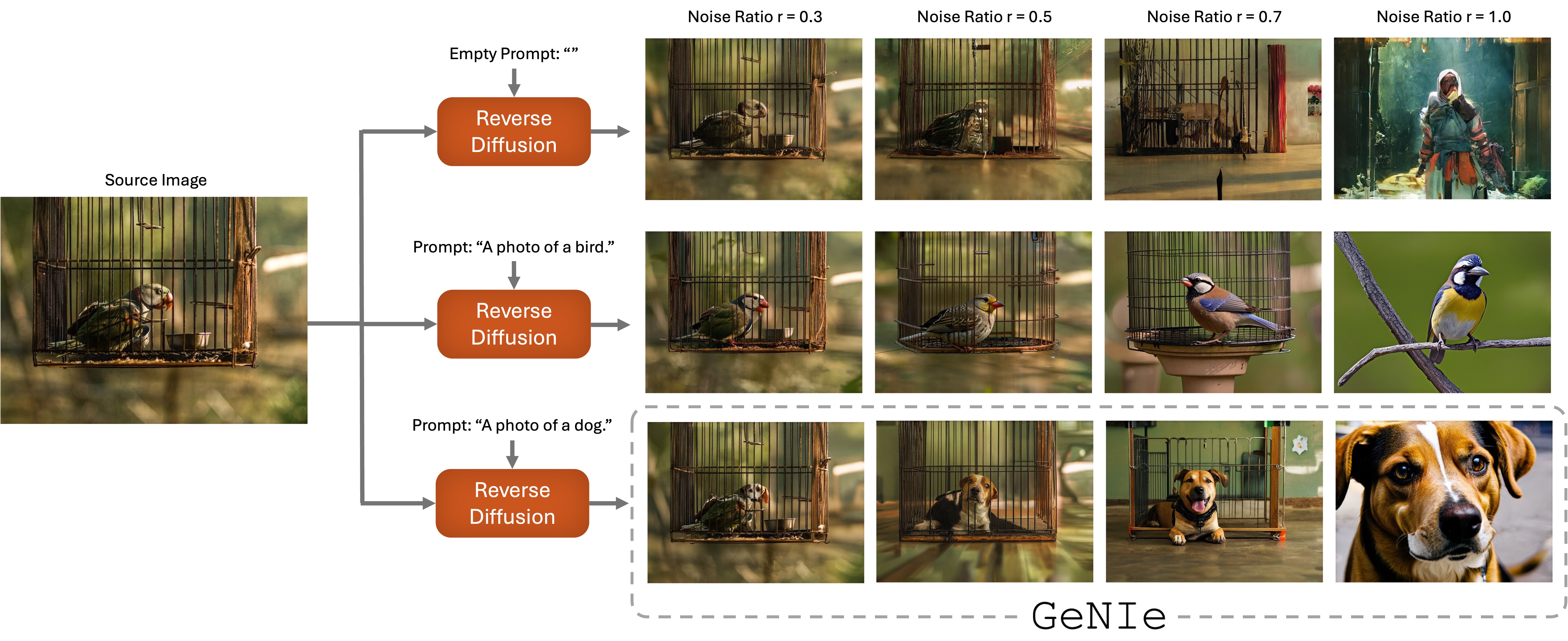}
\caption{\label{fig:genie5} {\bf Key components of \genie:} (i) careful choice of $r$ and (ii) contradictory prompt are two key idea behind \genie{}}
\end{figure}

% \subsection{Details of Fine-grained Dataset} 
% \label{sec:finegrain_appendix}
% We provide detailed information around fine-grained datasets in Table~\ref{tab:appendix_transfer_dset_details}, as discussed in Section~\ref{sec:finegrained}.

\subsection{Extra Computation of \genienoad{}}
Given that GeNIe-Ada searches for the best hard-negative between multiple noise-ratios $r$'s, it naturally requires a higher compute budget than $\texttt{txt2Img}$ that only uses $r=1$. For this experiment, we use GeNIe-Ada with $r \in \{0.6, 0.7, 0.8\}$ to compare with $\texttt{Txt2Img}$. Based on this, we only have $3$ paths, with steps of $0.1$), and for each of which we go through partial reverse diffusion process. E.g. for $r=0.6$ we do $30$ steps instead of standard $50$ steps of Stable Diffusion. This practically breaks down the total run-time of $\texttt{GeNIe-Ada}$ to approximately $2$ times that of the standard reverse diffusion (GeNIe-Ada: total $r = 0.6 + 0.7 + 0.8 = 2.1$ vs $\texttt{Txt2Img}$ total $r = 1$). Thus, to be fair, we generate twice as many $\texttt{Txt2Img}$ augmentations as compared to GeNIe-Ada to keep a constant compute budget across the methods, following your suggestion. The results are shown in Table~\ref{tab:ex_comput}. As can be seen, even in this new setting, GeNIe-Ada offers a performance improvement of $0.8\%$ to $1.9\%$ across different backbones.

\subsection{How does \genie{} control which features are retained or changed?}

We instruct the diffusion model to generate an image by combining the latent noise of the source image with the textual prompt of the target category. This combination is controlled by the amount of added noise and the number of reverse diffusion iterations. This approach aims to produce an image that aligns closely with the semantics of the target category while preserving the background and features from the source image that are unrelated to the target. 

To demonstrate this, in Figure~\ref{fig:genie5}, We are progessivley moving towards the two key components of GeNIe: (i) careful choice of $r$ and (ii) contradictory prompt. The input image is a bird in a cage. The top row shows a Stable Diffusion model, unprompted. As can be seen, such a model can generate anything (irrespective of the input image) with a large $r$. Now prompting the same model with ``a photo of a bird'' allows the model to preserve low-level and contextual features of the input image (up to $r = 0.7 $ and $0.8$), until for a large $r \geq 0.9$ it returns a bird but the context has nothing to do with the source input. This illustrates how a careful choice of $r$ can help preserve such low-level features, and is a key idea behind GeNIe. However, we also need a semantic switch to a different target class as shown in the last row where a hardly seen image of a dog in a cage is generated by a combination of a careful choice of $r$ and the contradictory prompt - leading to the full mechanics of GeNIe. This sample now serves as hard negative for the source image (bird class).

\subsection{More Visualizations} 
\begin{figure}[!b]
\centering

\includegraphics[width=1.0\linewidth]{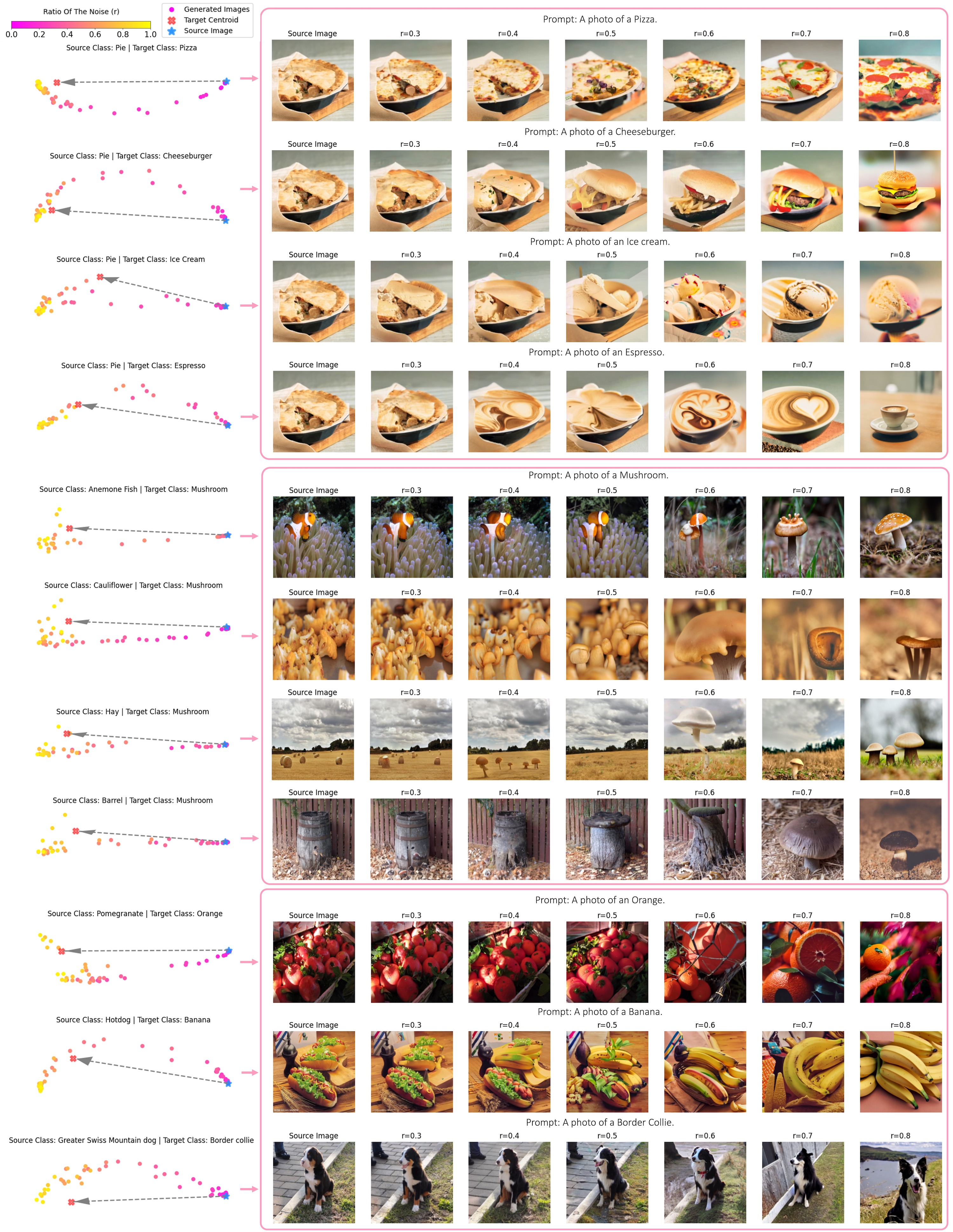}

\caption{\label{fig:traj_supp}{\bf Effect of noise in \genie:} Similar to Fig.~\ref{fig:traj}, we pass all the generated augmentations through the DinoV2 ViT-G model, which acts as our oracle model, to obtain their associated embeddings. Subsequently, we employ PCA for visualization purposes. The visualization reveals that the magnitude of semantic transformations is contingent upon both the source image and the specified target category.}

\end{figure}

\begin{figure}[t!]
\centering

\includegraphics[width=0.87\linewidth]{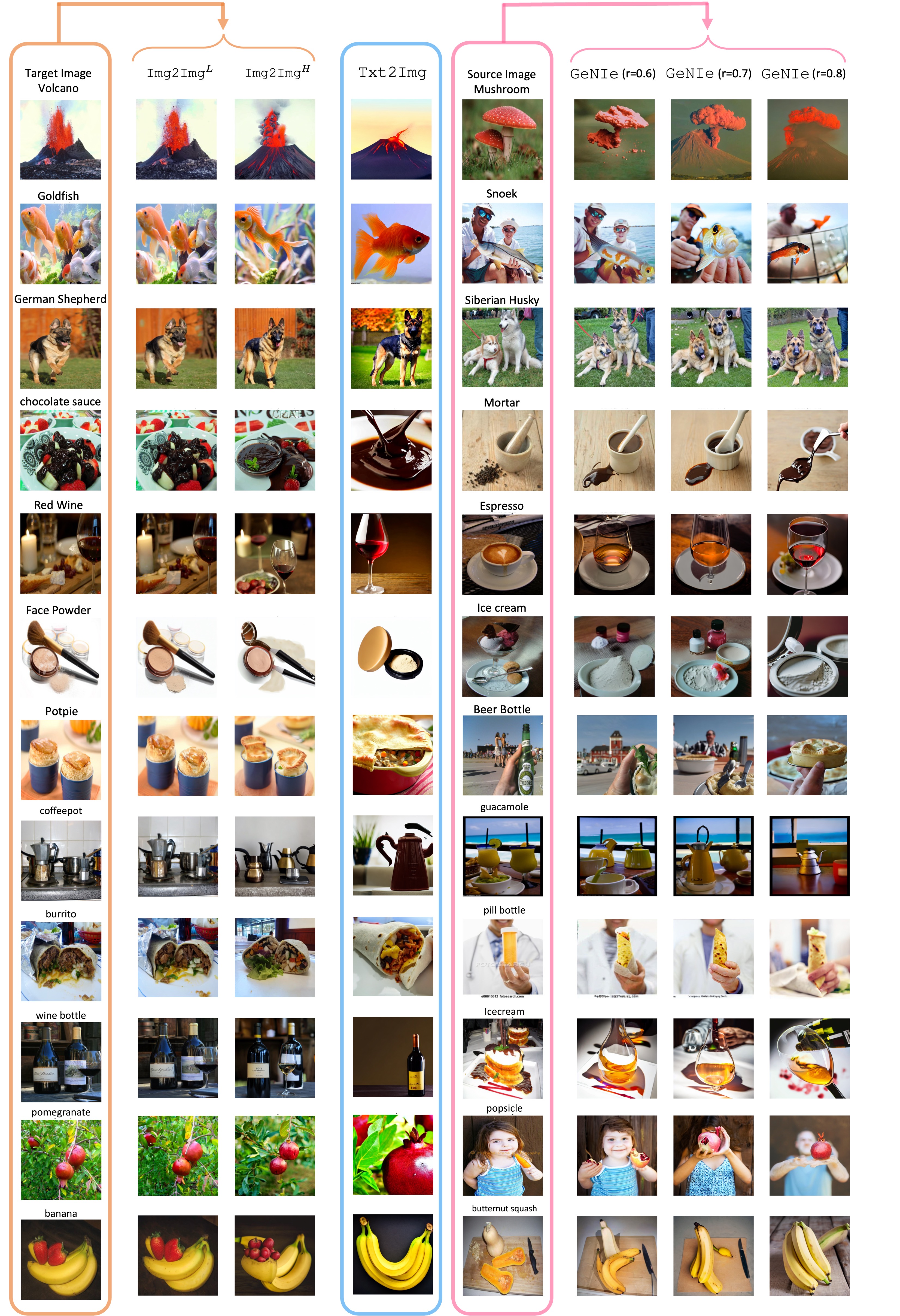}

\caption{{\bf Visualization of Generative Samples:} More visualization akin to Fig.~\ref{fig:vis1_genie}. We compare \genie{} with two baselines: \textbf{\posl{} augmentation} uses both image and text prompt from the same category, resulting in less challenging examples. \textbf{\txtimg{} augmentation} generates images based solely on a text prompt, potentially deviating from the task's visual domain. \textbf{\genie{} augmentation} incorporates the target category name in the text prompt along with the source image, producing desired images with an optimal amount of noise, and balancing the impact of the source image and text prompt. }
\label{fig:vis_genie_supp}

\end{figure}

\begin{figure}[t]
\centering

\includegraphics[width=1.0\linewidth]{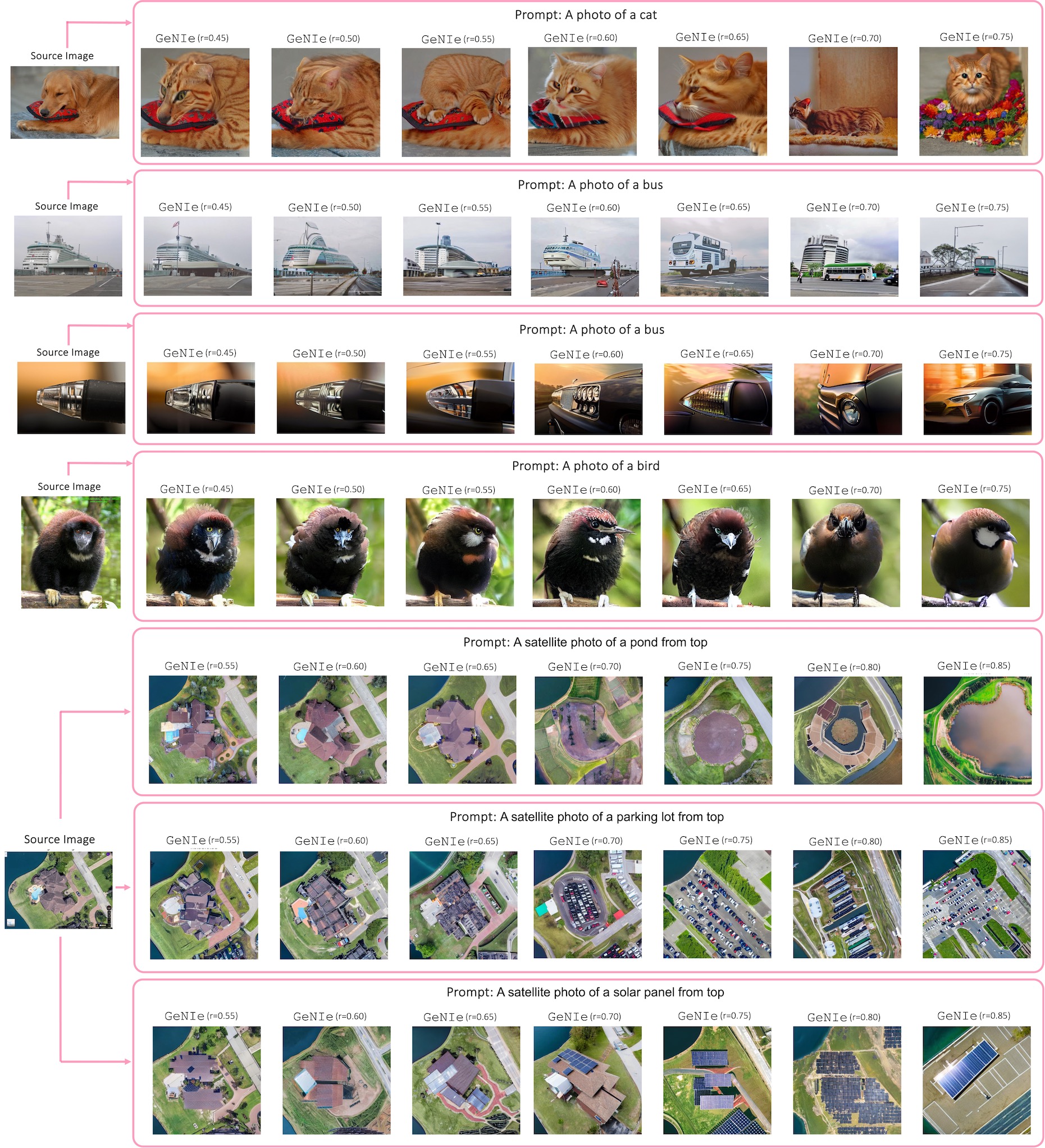}

\caption{\label{fig:noise_ab_supp}{\bf Effect of noise in \genie:} Akin to Fig.~\ref{fig:noise_ab}, we use \genie{} to create augmentations with varying noise levels. As is illustrated in the examples above, a reduced amount of noise leads to images closely mirroring the semantics of the source images, causing a misalignment with the intended target label.}

\end{figure}

Additional qualitative results resembling the style presented in Fig.~\ref{fig:vis1_genie} are presented in Fig.~\ref{fig:vis_genie_supp}, and more visuals akin to Fig.~\ref{fig:noise_ab} can be found in Fig.~\ref{fig:noise_ab_supp}. Moreover, we also present more visualization similar to the style in Fig.~\ref{fig:traj} in Fig.~\ref{fig:traj_supp}. 

\newpage
\clearpage

\end{document}